\documentclass[11pt]{article}

\usepackage[final]{acl}

\usepackage{times}
\usepackage{latexsym}

\usepackage{microtype}
\usepackage{hyperref}
\usepackage{url}
\usepackage{booktabs}
\usepackage{tabularx}
\usepackage{graphicx}
\usepackage{amsmath}
\usepackage{tcolorbox}
\usepackage{subcaption}
\usepackage{xcolor}
\usepackage{amssymb}
\usepackage{wrapfig}

\usepackage[T1]{fontenc}

\usepackage[utf8]{inputenc}

\usepackage{microtype}

\usepackage{inconsolata}

\usepackage{graphicx}
\usepackage[dvipsnames]{xcolor}

\newcommand{\metric}{\text{AUROC}}

\title{Testing the Limits of Truth Directions in LLMs}

\author{Angelos Poulis$^1$ \quad Mark Crovella$^{1,2}$ \quad Evimaria Terzi$^1$ \\
$^1$Department of Computer Science, Boston University \\
$^2$Faculty of Computing \& Data Sciences, Boston University \\
\texttt{\{apoulis,crovella,evimaria\}@bu.edu}
}

\begin{document}
\maketitle
\begin{abstract}
Large language models (LLMs) have been shown to encode truth of statements in their activation space along a linear truth direction.
Previous studies have argued that these directions are universal in certain aspects, while more recent work has questioned this conclusion drawing on limited generalization across some settings. In this work, we identify a number of limits of truth-direction universality that have not been previously understood.  
We first show that truth directions are highly layer-dependent, and that a full understanding of universality requires probing at \emph{many} layers in the model.  We then show that truth directions depend heavily on task type, emerging in earlier layers for \emph{factual} and later layers for \emph{reasoning} tasks; they also vary in performance across levels of task complexity. Finally, we show that \emph{model instructions} can affect truth directions; simple correctness evaluation instructions significantly affect the geometry and the generalization ability of the truth probes.
Our findings indicate that universality claims for truth directions are more limited than previously known, with significant differences observable for various model layers, task difficulties, task types, and prompt templates\footnote{Code and datasets: \url{https://github.com/angelosps/llm-truth-directions-limits}}.
\end{abstract}

\section{Introduction}
Large language models (LLMs) possess and manipulate vast amounts of knowledge, achieving remarkable performance on various tasks~\citep{bubeck2023sparksartificialgeneralintelligence}. However, their inner workings remain opaque, limiting the interpretability of their predictions. A growing body of work aims to understand these models by studying their internal representations. A key question in this research area is whether LLMs encode representations of truth. Such representations allow us to tell whether an LLM contains an internal representation of the correctness\footnote{Throughout the paper we use the terms ``truthfulness'' and ``correctness'' interchangeably.} -- either as decided by the LLM, when assessment requires some processing, or learned during pre-training -- of a statement. If so, then we say that there is a direction in the model's activation space that can separate statements based on their correctness. In this study, we extend the notion of truthfulness to statements that require processing beyond factual retrieval. While prior work focuses on probing for a truth direction separating facts based on their truthfulness, we also ask \emph{whether LLMs compute linear directions that separate computed results according to their correctness.}

To extract accurate and generalizable such directions, we need to answer several questions: \emph{where inside the LLM should we look for an accurate truth direction that will generalize the most across tasks?} \emph{Does instructing the model to assess correctness affect the geometry of truth probes?} and, \emph{What limitations prevent decoding strong truth directions?}

Previous studies \citep{marks_tegmark_2024,burger_2024,bao-etal-2025-probing} have analyzed truth directions at specific layers, without instructing the model, and without controlling for task difficulty, leaving open the question of how these choices affect conclusions about universality. In this work, we examine the extent to which linear truth directions in LLMs are universal by systematically varying three dimensions: (i) the layer at which we probe for truth directions, (ii) the introduction of model instructions, and (iii) the difficulty of the task. We introduce a suite of controlled datasets spanning factual knowledge and arithmetic reasoning, where task difficulty is operationalized by the number of reasoning steps required to assess correctness. To the best of our knowledge, we are the first to conduct such a detailed and systematic study of linear truth directions.

We find that linear truth directions in LLMs are reliable \emph{primarily in factual recall cases}, and break down when truth assessment of statements depends on computing and storing intermediate results. Further, we show that both the choice of layer for probing such directions, and whether we instruct the model to assess correctness, significantly influence truth directions. Conclusions about truth directions are highly layer-dependent; activations at early layers appear to capture surface-level features that may confound truth extraction, resulting in probes with limited generalization. Moreover, we find that explicitly asking the model to assess correctness, significantly affects the geometry of the learned truth directions, and can improve generalization of probes to factual retrieval tasks.
\section{Related work}

\paragraph{Truth directions in LLMs.}
Early work on truth directions in LLMs demonstrated that unsupervised~\citep{burns2023discovering}, and supervised  linear~\citep{inferencetimeintervention} and non-linear classifiers~\citep{azaria2023internalstatellmknows} trained on hidden states of LLMs can separate activations that correspond to true versus false statements. However, \citet{Levinstein_2024} found that the classifiers of \citet{azaria2023internalstatellmknows} fail to generalize to negated versions of these statements. 
\citet{marks_tegmark_2024} presented evidence that truth representations in LLMs emerge linearly for true/false datasets. Specifically, they probed hidden states identified as causally influential and showed that truth directions generalize across topics and sentence structures. However, their analysis was conducted at specific layers, with no model instructions, and exclusively on factual datasets. We extend these studies by systematically probing all layers of the model, using model instructions, and introducing factual and arithmetic tasks that require multi-step reasoning. \citet{burger_2024} investigated why probes trained only on affirmative statements fail to generalize to negated variants. In particular, they proposed a two-dimensional subspace along which truth is linearly decoded irrespective of the sentence polarity. Their analysis focused on a single model layer. In contrast, we show that their finding is specific to the selected layer, which happens to reflect a transitional phase in the model's computation of truth. \citet{savcisens2026trilemmatruthlargelanguage} challenged the binary-probe setup and proposed multi-class classification using bag-of-token representations. Their work tested generalization across knowledge domains, whereas we focus on how task difficulty affects generalization within a domain using statements based on objective truth. \citet{schouten2025truthvalue} showed that truth directions are context-sensitive: adding a premise before the probed hypothesis can shift the probe's truth assessment. We ask a related but different question: whether explicit evaluation instructions for truth assessment can affect the geometry and generalization of truth probes. Our setup does not introduce additional information in-context; instead, it prompts the model to assess the correctness of the input statement.

\paragraph{Generalization of truth directions.}
\citet{liu-etal-2024-universal} found that single-task probes fail to generalize across tasks but generalization improves when training on over 40 diverse datasets. Our study characterizes the generalization failure of single-task probes along layers, model instructions and task difficulty.

\citet{bao-etal-2025-probing} tested how truth directions generalize across simple logical transformations of statements, and to complex datasets such as MMLU/TriviaQA, where they found strong generalization for the former but weaker for the latter. However, the reason for this remains unclear: it could be the diverse knowledge domain these benchmarks test, possible question ambiguity, or task complexity. 

We focus on task complexity, by systematically increasing the number of steps required to assess correctness. We find that the generalization ability of truth probes degrades significantly with a few such steps, indicating that the computational demand of the task is a limiting factor for strong truth directions.
\citet{haller2025llmknowledgebrittletruthfulness} studied the robustness of truth probes to out-of-distribution perturbations. 
Their work tested robustness by corrupting the input statements, whereas we increase the number of steps required to decide truth. 
Both approaches find that truth probes are more fragile than previously assumed, but our setup isolates the effect of task difficulty.
\citet{ying2026truthfulnessspectrumhypothesis} proposed that LLMs encode truth along a spectrum of generality, identifying the type of truth (empirical, logical, fictional) as a boundary for generalization. Their work varied the type of truth across tasks of similar difficulty, whereas we vary the difficulty of tasks within fixed domains.

\paragraph{Input-truth vs.\ output-truth.} \citet{liu-etal-2023-cognitive} documented three cases where truth probes disagree with model outputs, finding that probes are often better calibrated on uncertain outputs. A parallel line of work studies directions in LLMs that capture the correctness of models' outputs~\citep{orgad2025llms,azizian2025geometriestruthorthogonaltasks,cencerrado2026answerneededpredictingllm,zhang2025reasoningmodelsknowtheyre, long-etal-2025-truthful}. This is different from our study and those cited above, which focus on correctness of the input statements. 

\section{Experimental setup}\label{sec:experimental_setup}
\paragraph{Datasets.} We hypothesize that LLMs represent correctness of arithmetic expressions differently from factual statements. To investigate this, we probe truth directions across \emph{factual tasks} that require retrieving facts from the model, and \emph{arithmetic tasks} that require performing numerical computations.
For factual tasks, we build on publicly available datasets used in previous papers. Specifically, we use the \texttt{cities, neg\_cities, cities\_conj} datasets from~\citet{marks_tegmark_2024}, which contain unambiguous sentences of the form ``The city of Paris is in France,'' negated variants of those, and the conjunction of two such sentences respectively. Each statement is associated with a binary label indicating its ground truth. We extend these base datasets to create six factual tasks (F0--F5) across two groups: factual retrieval (F0--F2), and factual counting (F3--F5). Each task within the group increases in complexity by varying the number of independent factual retrievals required, and the operations needed to combine them (negation, conjunction, and counting constraints).
For arithmetic tasks, we create a set of three synthetic datasets (A1--A3) consisting of equations with one, two, or three operations from the set $\{+,-,/,\times\}$ over integers in the range $(0,100)$. Each equation is paired with a proposed answer and labeled as true or false.
Table~\ref{tab:tasks_difficulty} summarizes all tasks;
details on dataset generation can be found in Appendix~\ref{app:dataset_creation}.
\paragraph{Operationalizing truth.} Assessing the truthfulness of a given statement can require a diverse set of mechanisms depending on the task, as in the retrieval and computation cases above. By operationalizing truth uniformly across tasks as agreement with the gold label of the statement, we can test whether the model has a universal, linearly decodable truth direction across these cases.

\paragraph{Characterizing task difficulty.} To characterize task difficulty, we use the number of discrete operations required to verify correctness of the input.

For arithmetic tasks, difficulty corresponds to the number of arithmetic operators (except the equal sign ``$=$'') in the expression: A1 contains expressions of one operator (e.g., $3\times 12 = 36$), A2 of two operators (e.g., $(9+3)\times6=72$), and A3 of three operators (e.g., $(2\times5) - (9/3) =7$). Each additional operator requires the model to compute and maintain intermediate results to decide correctness of the provided expression. 

For factual tasks, difficulty is characterized by the number of factual retrievals from parametric memory, and the operations needed to combine them. All factual tasks are operating within the same knowledge base (e.g., knowledge of city locations), thus processing demand of a task is independent of knowledge availability in the model. 

Together, these yield an intuitive (rather than a strict) ordering of the tasks \textit{within} each group that enables the study of truth directions along the difficulty axis.

\begin{table}[t]
    \centering
    \tiny
    \caption{Tasks with increasing reasoning steps within each group.}
    \label{tab:tasks_difficulty}
    \begin{tabularx}{.47\textwidth}{c c X}
        \toprule
        \textbf{Task} & \textbf{Operations} & \textbf{Example statement} \\
        \midrule
        \multicolumn{3}{@{}l}{\textbf{Factual retrieval}} \\
        F0 & 1 fact &
        ``The city of Paris is in France.'' \\
        F1 & 1 fact + negation &
        ``The city of Paris is not in Australia.'' \\
        F2 & 2 facts + conjunction &
        ``It is the case both that the city of Paris is in France and
        the city of Dortmund is in Germany.'' \\
        \multicolumn{3}{@{}l}{\textbf{Factual counting}} \\
        F3 & counting over 2 cities$^\dagger$ &
        ``Exactly 2 of the following cities are in France: Paris,
        Marseille.'' \\
        F4 & counting over 5 cities &
        ``Exactly 2 of the following cities are in Germany: Athens,
        Paris, Munich, Beijing, Dortmund.'' \\
        F5 & double counting over 6 cities &
        ``Exactly 2 of the following cities are in Germany and 1 in
        Greece: Athens, Paris, Munich, Beijing, Berlin, Boston.'' \\
        \midrule
        \multicolumn{3}{@{}l}{\textbf{Arithmetic}} \\
        A1 & 1 arithmetic op &
        ``$37 + 15 = 52$.'' \\
        A2 & 2 arithmetic ops &
        ``$(37 + 15) - 4 = 48$.'' \\
        A3 & 3 arithmetic ops &
        ``$(37 + 15) / (12 - 8) = 13$.'' \\
        \bottomrule
    \end{tabularx}
    \smallskip\par\noindent
    {\footnotesize $^\dagger$~F3 can be solved with a shortcut, by direct comparison (analogous to F2); it is included to test whether models exploit this heuristic rather than perform genuine counting.}
\end{table}

\paragraph{Models.} We probe four models, from two different families: Llama~\citep{grattafiori2024llama3herdmodels}, and Gemma~\citep{gemma2}. Specifically, we use Llama-3.2-3B-Instruct, Llama-3.1-8B-Instruct, Gemma-2-2b-it, and  Gemma-2-9b-it. In the main paper, we focus on Llama-3.1-8B-Instruct, which has 32 layers and a residual stream dimension $d_{\text{model}}=4096$. We provide details of these models in Appendix~\ref{app:sec:model_details}.  The results for all models are presented in Appendix~\ref{app:sec:other_model_results}.

\paragraph{Probes.} Prior work identifies truth directions in model activation space using probing, i.e., training a classifier in a supervised manner on a labeled set of model activations~\citep{alain2016, belinkov-2022-probing}. Since prior work supports that truth of statements can be \emph{linearly} decoded from LLM activations~\citep{inferencetimeintervention, marks_tegmark_2024}, we use linear probes in our study. A linear probe is defined as follows: 
let $h_i^{(\ell)}\in\mathbb{R}^d$ denote the representation of example $i$ at layer $\ell$ of the model, and let $y_i\in\{0,1\}$ be its binary label. For each layer $\ell$, we compute the training-set mean $\mu^{(\ell)}=1/N\sum_{i=1}^N h_i^{(\ell)}$,
and define mean-centered representations $\tilde{h}_i^{(\ell)} = h_i^{(\ell)}-\mu^{(\ell)}.$
We then train a bias-free logistic linear probe:
\[
p(y_i=1 \mid \tilde{h}_i^{(\ell)})=\sigma\!\left({w^{(\ell)}}^\top \tilde{h}_i^{(\ell)}\right),
\]
where $w^{(\ell)}\in\mathbb{R}^d$ is the probe vector and $\sigma(z)=\frac{1}{1+e^{-z}}$ is the sigmoid function. Thus, we define $w^{(\ell)}$ as the \emph{truth direction} at layer $\ell$. Probe training details can be found in  Appendix~\ref{app:sec:probe_training}.

A key question when probing for truth directions is \emph{where} in the transformer architecture to extract representations. In particular, we need to decide which \emph{layer} of the model and which \emph{token} position within that layer to probe. Following prior work~\citep{marks_tegmark_2024, burger_2024, bao-etal-2025-probing}, we extract activations from the residual stream at the \textit{final} token position, as the causal attention mechanism ensures this position has aggregated information from the entire input sequence\footnote{All statements in our datasets end with a final period (\texttt{.}) token.}. We extract model activations using the TransformerLens library \citep{nanda2022transformerlens}.
\paragraph{Evaluation methods.} 
We evaluate the performance of a probe by measuring the area under the receiver operating characteristic curve ($\metric$) metric.
To understand the universality of truth directions, we use two main methods: (a) the comparison of truth directions \emph{geometrically}, and (b) the \emph{generalization} of truth directions across tasks. For the former, we use the cosine similarity between two truth directions; for the latter we measure the $\metric$ score of a probe trained on one task when applied to another task.

\section{Truth directions across layers}\label{sec:layers}
The first question we address concerns the layer at which we can extract accurate truth directions. Previous work on truth directions typically fixes a particular layer for probing model activations. The specific layer has been chosen using activation patching~\citep{marks_tegmark_2024}, or by calculating the ratio of \emph{between- to within-class variance of activations}~\citep{burger_2024,bao-etal-2025-probing}. However, if truth directions change across the model's depth, single-layer analyses may capture layer-specific phenomena. When calculating the between- to within-class variance ratio for our tasks across layers (Figure~\ref{fig:app_variance_ratio}, App. \ref{app:layer-selection}), we observe that different tasks peak at different layers. Thus, no single layer is universally optimal for probing truth directions, and we need to probe all layers in order to understand the complete picture of truth directions in the model. 

\begin{figure*}[t]
\centering
\begin{minipage}[t]{0.41\textwidth}
    \vspace{0pt}
    \subfloat[In-domain test AUROC across layers.\label{fig:test_auroc_per_layer}]{%
        \includegraphics[width=\linewidth]{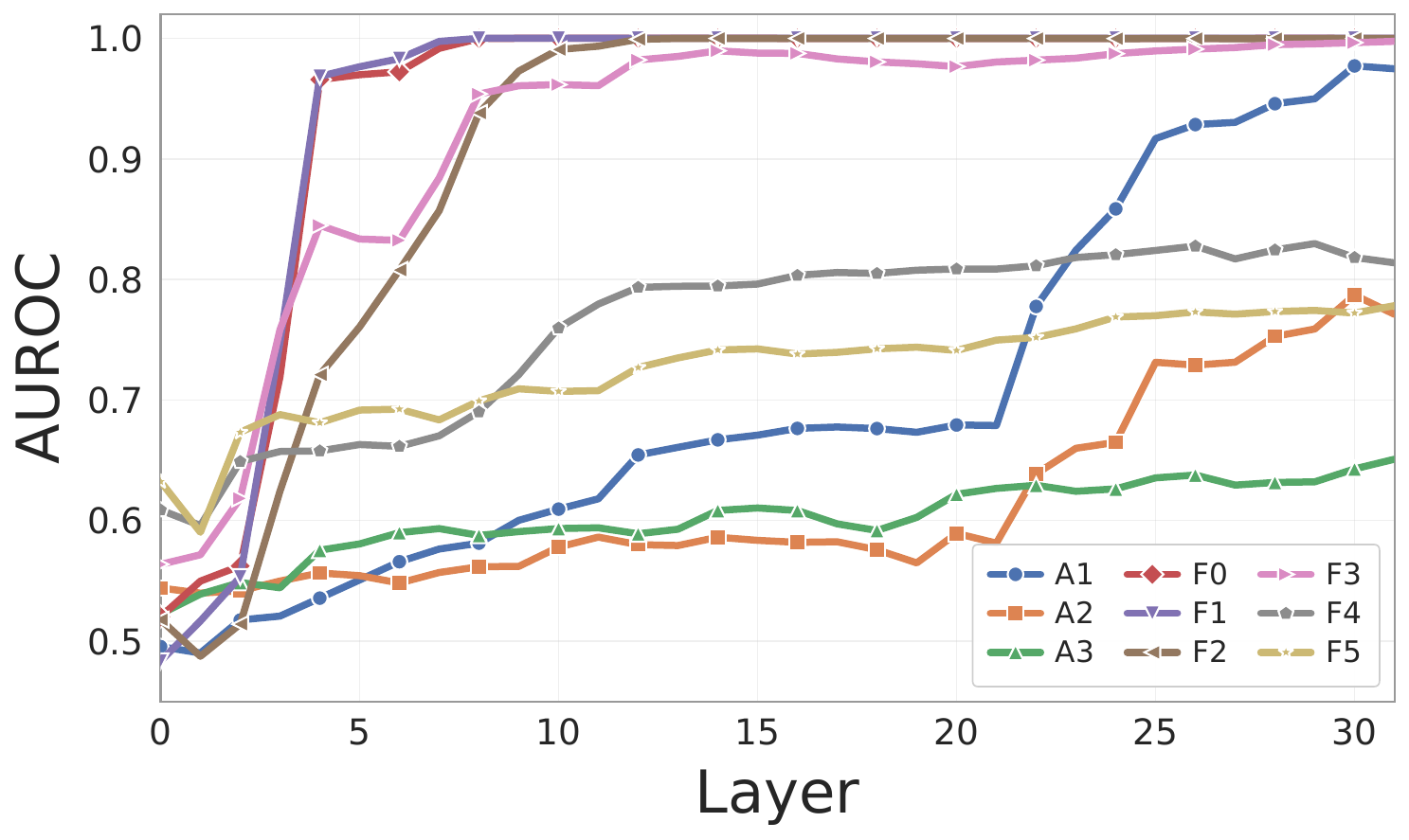}%
    }\\[0.5em]
    \subfloat[Cross-task generalization AUROC of the F0-trained probe across layers.\label{fig:f0_generalization_per_layer}]{%
        \includegraphics[width=\linewidth]{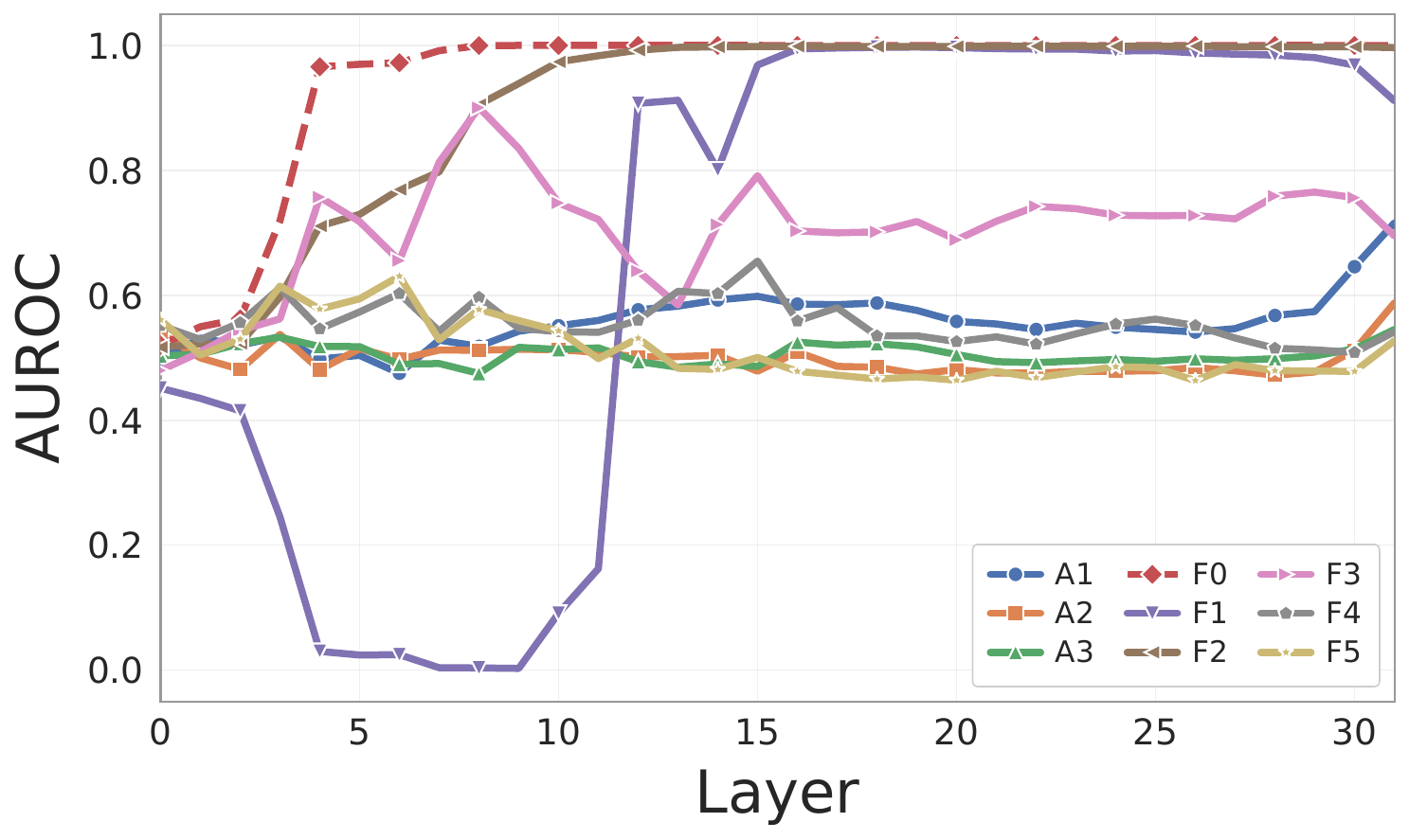}%
    }
\end{minipage}%
\hfill
\begin{minipage}[t]{0.56\textwidth}
    \vspace{0pt}
    \subfloat[Cosine similarity of probes across all pairs of layers for each task.\label{fig:probe_cos_per_layer}]{%
        \includegraphics[width=\linewidth]{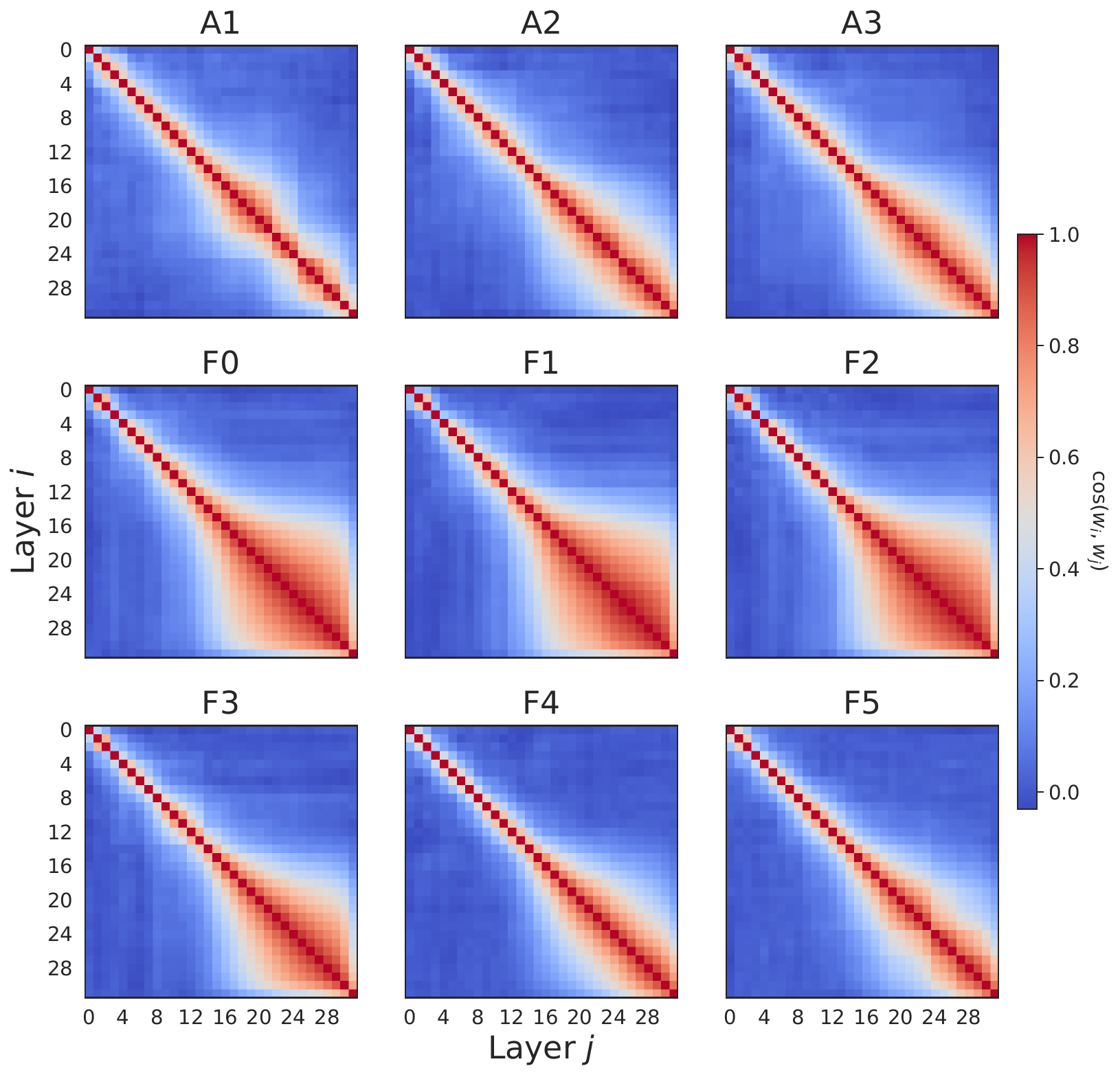}%
    }
\end{minipage}
\caption{Layer dependence of truth directions. (a) In-domain test-set performance. (b) F0-probe cross-task generalization. (c) Cosine similarity of probes across layer pairs.}
\label{fig:layer_dependence_combined}
\end{figure*}


\subsection{Truth directions peak early for factual tasks, later for arithmetic}
We train a linear probe at each layer and evaluate on the held-out test set of the same task. Figure~\ref{fig:test_auroc_per_layer} shows three patterns of truth direction emergence: tasks F0--F3 reach almost-perfect performance in early to mid layers, arithmetic tasks (A1--A3) emerge much later, and factual counting tasks (F4--F5) emerge late, similar to arithmetic tasks. In factual retrieval tasks, the truth of a statement can be determined using a simple retrieval from parametric knowledge (F0, F1), or by combining two factual retrievals (F2). On the other hand, arithmetic tasks require computing the result of expressions to assess correctness, which as recent research suggests is happening in mid-late layers of LLMs~\citep{nikankin2025arithmeticalgorithmslanguagemodels}, consistent with the finding that more complex concepts require deeper layers to emerge~\citep{concept-depth}. Although F3 structurally requires counting over two cities, it can be solved by direct comparison of two independent facts, similarly to the F2 task. We observe that the truth direction emergence for F3 follows the pattern of F0--F2, suggesting that the model exploits this shortcut rather than performing genuine counting. 

Tasks F4--F5 are extensions of the F3 task, but they require counting over a list of more than two cities, hence the shortcut above cannot be used in this case. Keeping track of such counters for truth evaluation appears to cause truth directions to emerge later compared to tasks that can be solved without the need for genuine counting (F0--F3). 

\subsection{Layer choice affects generalization of truth directions}\label{sec:layer_generalization}
We investigate whether the cross-task generalization of truth probes varies across layers. For simple factual tasks (F0--F3), probes achieve high in-domain $\metric$ from early layers, which suggests that any such layer is equally appropriate for extracting a truth direction. However, we find that \emph{generalization behavior varies across layers even when in-domain performance is very high}. To isolate this, we train probes on a single-source task at each layer and evaluate them on all other tasks at the same layer.
Figure~\ref{fig:f0_generalization_per_layer} shows cross-task generalization for probes trained on F0, the simplest factual task. 
Although F0-trained probes achieve near-perfect performance from layer 4, they fail on F1, the negated variant of F0, at that layer. In particular, F0-trained probes in layers 4--10 show \emph{inverted} separation on F1 ($\metric \approx 0$), systematically misclassifying true statements as false and vice versa. This generalization failure matches prior work observation~\citep{Levinstein_2024, burger_2024} that probes trained on affirmative statements (i.e., sentences without ``not'') fail to classify their negated variants. This resolves in mid layers where F0 generalizes well to F1, indicating that \textbf{early layer probes may capture surface features rather than truth.}

To understand why this happens, we measure the fraction of truth-related variance explained by two directions: a polarity-dependent ($t_p$) and a polarity-invariant ($t_G$) direction, following the setup of \citet{burger_2024}. The polarity direction classifies effectively affirmative statements but classifies oppositely their negated variants, whereas the general truth direction effectively classifies truth irrespective of sentence polarity. In Figure~\ref{fig:app_t_g_t_p_variance}, we show the fraction of truth-related variance these two directions explain across model layers. 
In early layers, $t_p$ dominates ($\sim0.38$ at layer 7, versus $\sim0.1$ for $t_G$), confirming that \emph{early-layer truth probes primarily capture sentence polarity}, and that the truth direction has not been computed by the model yet. At layer 12, the layer analyzed by \citet{burger_2024}, $t_p$ and $t_G$ explain similar fractions of truth-related variance ($\sim0.33$ each). However, by the middle layers, $t_G$ takes over and $t_p$ decays, indicating that the truth-direction disentangles from the polarity direction as the processing evolves. Therefore, the two-dimensional subspace reported by~\citet{burger_2024} seems to reflect a stage of transition in the model's processing, rather than a universal property of truth directions.

Figure~\ref{fig:probe_cos_per_layer} shows the geometric transition of truth directions across layers. For the factual tasks F0--F3, probes show a sharp transition in middle layers: while probes from early layers point in different directions, late-layer probes are more similar to each other. \textbf{This indicates that the model is converging to a more stable truth direction} (as supported by generalization performance in Figure~\ref{fig:f0_generalization_per_layer}). In contrast, the probes for the tasks F4--F5 and arithmetic tasks A1--A3 show a weaker convergence pattern over layers.


These results imply that (i) \textbf{the depth at which truth becomes linearly separable depends on whether truth can be determined solely using factual retrieval or requires multi-step reasoning}, and (ii) \textbf{single-layer analyses can be misleading}, since early-layer truth directions may reflect surface features (e.g., sentence polarity) with limited cross-task generalization. 
\section{Model instructions change the geometry of truth probes}\label{sec:task_framing}

We hypothesize that explicitly instructing the model to evaluate the correctness of the given statement may change the geometry of truth directions. This is an example of how the choice of prompt can significantly affect the performance of an LLM on a task~\citep{gonen-etal-2023-demystifying}, motivating the question of \emph{``what is the effect of model instructions on truth directions?''} \citet{schouten2025truthvalue} studied how in-context premises affect truth directions. Here, we focus on \emph{instructing the model to assess correctness without additional in-context information.} To do so, we compare probes obtained under two prompt templates:

\begin{tcolorbox}[
    colback=Maroon!7,
    colframe=Maroon,
    colbacktitle=Maroon,
    coltitle=white,
    fonttitle=\bfseries,
    title=Prompt templates for model instructions,
    rounded corners,
    boxrule=1pt,
    left=6pt, right=6pt, top=4pt, bottom=4pt
]
\textbf{Passive (no-prompt):}~\texttt{The city of Paris is in France.}\\[4pt]
\textbf{Explicit evaluation (ask-correct):}~\texttt{Is the following correct?\,${\scriptstyle\hookleftarrow}$\,The city of Paris is in France. Answer:}
\end{tcolorbox}

We focus on these two templates since they represent a clear contrast: passive statement processing (by simply presenting a statement without any additional token) versus active correctness assessment by asking the model whether the provided statement is correct before presenting the statement. The two templates differ in their final, and hence the one being probed, token (``\texttt{.}'' in no-prompt and ``\texttt{:}'' in ask-correct). To verify this does not confound our analysis, we run a control experiment replacing the colon with a period in the ask-correct template and find that our results are consistent with either last token in the layers where truth directions emerge (see Appendix~\ref{app:token_control}). Here, we keep the colon version since it is a more natural evaluation template. We also report results for additional template variations in Appendix~\ref{app:sec:other_prompts_results}.

\subsection{Truth directions are not universal across prompt templates} 

We test whether truth directions are aligned across prompt templates using both geometric and generalization evaluations (see Section~\ref{sec:experimental_setup}).

\begin{figure}
    \centering
    \includegraphics[width=\linewidth]{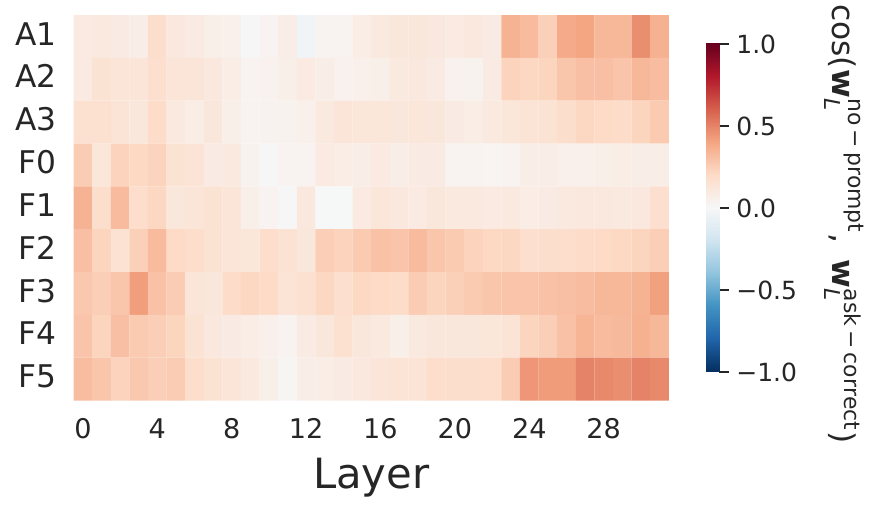}
    \caption{Cosine similarity between no-prompt and ask-correct probes per layer.}
    \label{fig:cosine_no-prompt_ask-correct}
\end{figure}

\paragraph{Geometric evaluation.} We compute the cosine similarity between probes trained on the same task under different prompt templates across layers. Figure~\ref{fig:cosine_no-prompt_ask-correct} shows that the probes trained under no-prompt have largely low cosine similarity with the probes trained on ask-correct activations. Specifically, the directions have low cosine similarity across tasks in early and middle layers; for the factual retrieval tasks (F0--F2) probes become almost orthogonal with model instructions. In contrast, the tasks A1--A3 and F3--F5 show higher similarity in the last layers. Interestingly, probes trained under different model instructions are highly aligned with each other (see detailed heatmaps in Figure~\ref{fig:app_cosine_all_pairs}), despite different wording of the instructions. This suggests that model instructions in the prompts significantly influence the geometry of the learned probe directions. 
\begin{figure*}[ht]
    \centering
    \includegraphics[width=\linewidth]{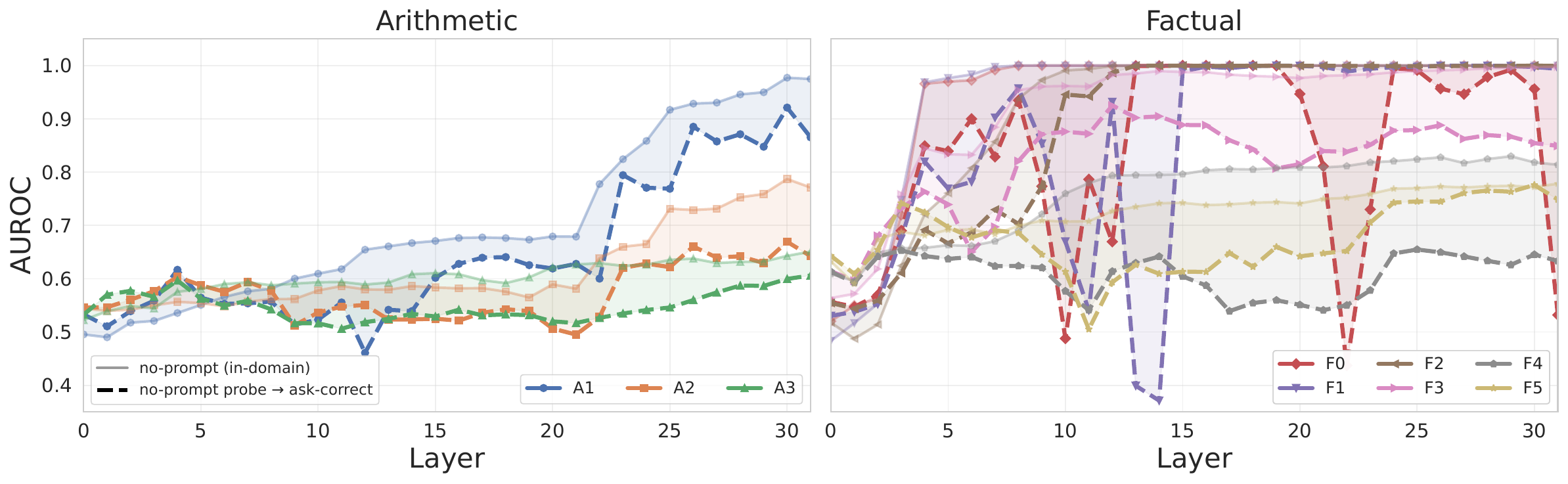}
    \caption{Probes trained on no-prompt activations do not generalize consistently to ask-correct activations.
    Left: probes for arithmetic tasks are more robust than factual tasks to this template.
    Right: factual tasks show significant performance variations under the ask-correct instructions over layers.}\label{fig:row1_to_row2d_generalization}
\end{figure*}

\paragraph{Generalization evaluation.} Figure~\ref{fig:row1_to_row2d_generalization} shows the performance drop when no-prompt probes are evaluated on their own test-set versus on ask-correct activations. For arithmetic tasks, generalization is degraded on ask-correct activations, but does not collapse: 
the largest drop is observed for A1 in layer 12, where the in-domain performance is $\metric\approx 0.65$ 
and the truth probe has not converged to its peak performance.
For factual tasks, model instructions cause sharp drops in generalization in specific layers for F0 and F1, while F3--F5 consistently show a large generalization gap between the two prompts.

Together, these results show that truth probes are not invariant to model instructions; even a simple correctness evaluation instruction in the prompt can affect the geometry and the generalization ability of the learned truth directions. 


\subsection{Instructing for truth assessment changes cross-task generalization}

\begin{figure*}[ht]
    \centering
    \includegraphics[width=\linewidth]{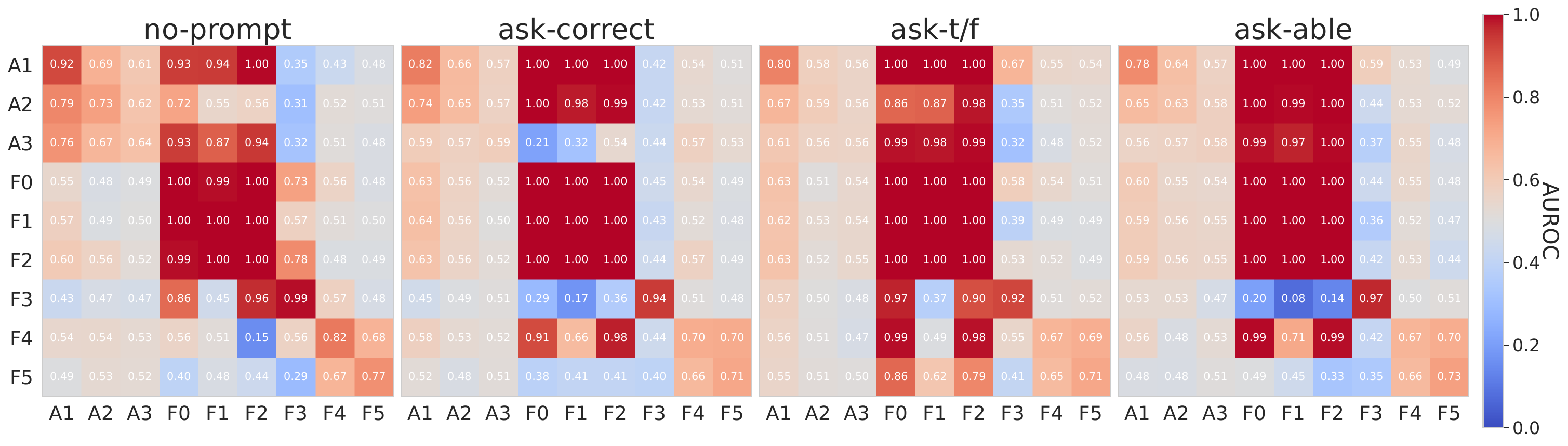}
    \caption{
    Effect of model instructions on generalization of truth directions.
    Rows correspond to the task used for probe training and columns to the task used for evaluation. Probes trained and evaluated on activations from layer 25.}
    \label{fig:prompt_generalization_heatmap}
\end{figure*}

Given that model instructions change the geometry of the probe directions, we ask \emph{whether the cross-task generalization of probes is also affected}. We test this using the generalization evaluation method outlined in Section~\ref{sec:experimental_setup}.

Figure~\ref{fig:prompt_generalization_heatmap} shows the cross-task generalization of truth directions across four templates (see Appendix~\ref{app:prompts} for details).
Under no-prompt (Figure~\ref{fig:prompt_generalization_heatmap} left), the arithmetic-trained probes partially generalize to the factual retrieval tasks F0--F2: the A1-trained probe generalizes very well (mean-AUROC $=0.96$ at layer 25), but A2 shows weak generalization. The pattern is different for prompts with instructions (Figure~\ref{fig:prompt_generalization_heatmap} three rightmost heatmaps) where higher generalization scores are more concentrated in F0--F2 tasks compared to no-prompt where higher scores seem to be more spread out. Further, the generalization of arithmetic probes to F0--F2 is noticeably improved compared to the no-prompt setting: mean-$\metric$ of A1$\rightarrow$F0--F2=1.0, A2$\rightarrow$F0--F2=0.96 across all instructed prompts. The same pattern is observed for A3-trained probes across instruction prompts, except ask-correct where also the in-domain test-set performance of the probe is $\metric=0.59$.


To confirm that these effects are not caused by confounding variables (e.g., extra tokens in the prompt), we test two control prompt templates in Appendix~\ref{app:control_experiment}. 

Although explicit evaluation instructions may partially align truth directions across tasks and improve generalization in some cases, truth directions fail to generalize to harder tasks because the activations themselves remain highly entangled (see Section~\ref{sec:difficulty}).

\section{Truth directions across tasks of increasing difficulty}\label{sec:difficulty}

Finally, we examine the effect of task difficulty by varying the number of intermediate steps required to assess correctness. 
\textbf{The controlled design of our set of tasks allows us to test how far from pure factual recall a task can be before the generalization ability of truth directions degrades.} 


\begin{figure}[t]
    \centering
    \includegraphics[width=\linewidth]{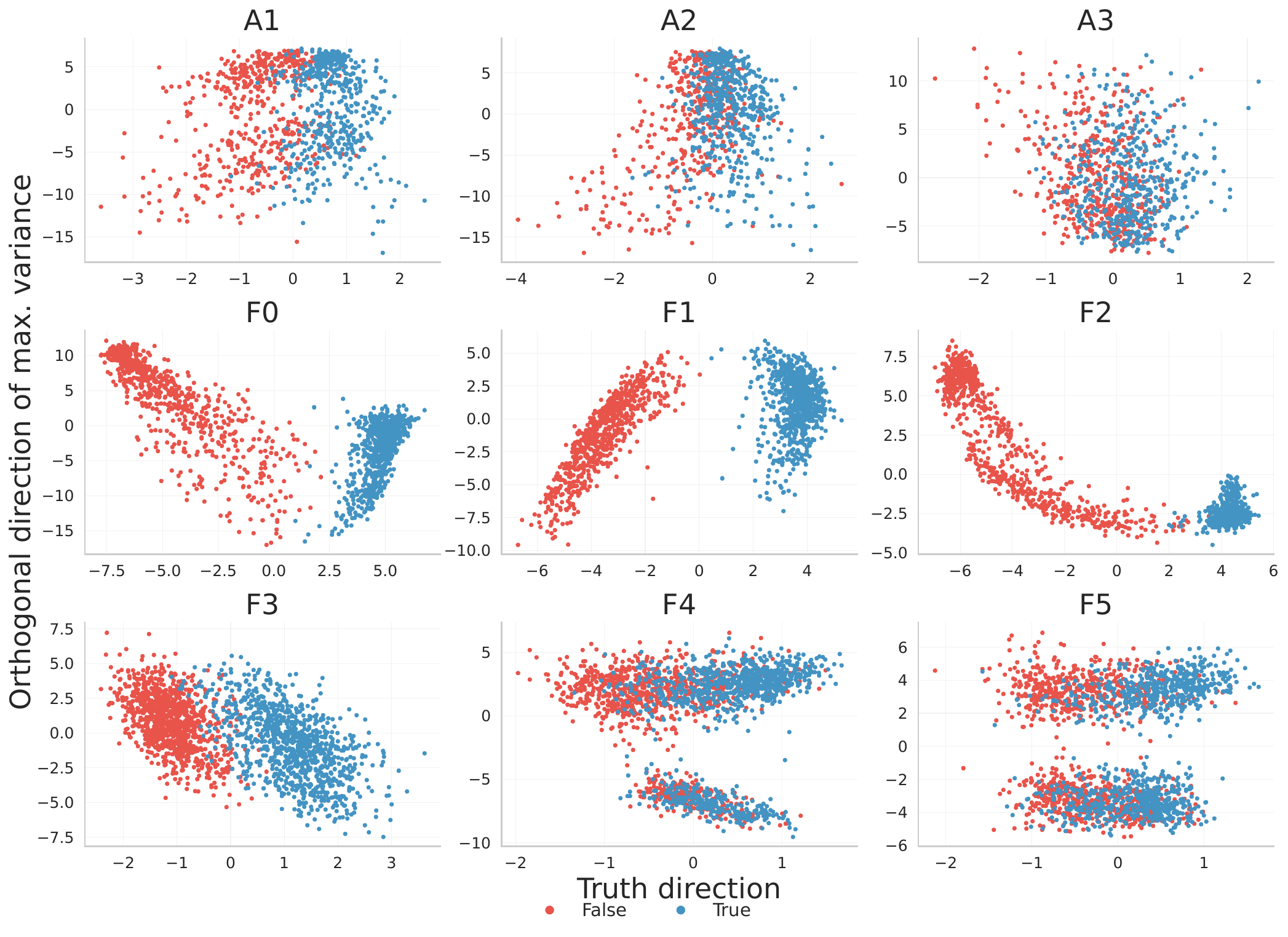}
    \caption{Two-dimensional projections of model activations extracted from layer 25 on the learned truth direction and on the direction of maximum variance in the orthogonal space.}
    \label{fig:truth_difficulty_scatter}
\end{figure}

\paragraph{Probes fail to generalize on factual counting tasks.} 
Figure~\ref{fig:prompt_generalization_heatmap} shows that truth directions trained on F0 generalize well to the factual retrieval tasks F1--F2, but generalization drops significantly on the factual counting tasks except the F3 which can be solved similarly to F2 as discussed in Section~\ref{sec:experimental_setup}. Strikingly, training a probe on a more complex source task does not resolve this: an F3-trained probe achieves $\metric=0.57$ on F4, below the F4-trained probe's performance ($\metric = 0.82$). Since truth is linearly decodable in F4, the generalization failure cannot be attributed to the model's lack of capability to perform the task. We locate the boundary at which the in-domain performance of truth probes degrades, to counting over lists of length 3 (see Appendix~\ref{app:intermediate_difficulty}).


\paragraph{Probes fail to generalize on multi-step arithmetic tasks.} 
Figure~\ref{fig:prompt_generalization_heatmap}  shows a similar pattern for arithmetic tasks, where truth directions decay rapidly beyond single-operation expressions. Specifically, probes trained on A1 show strong test-set performance but degrade significantly when evaluated on A2, and even more on A3. Again, training a probe on a more complex source task (A2) does not resolve this ($\metric < 0.65$ on A3).

\paragraph{Geometric evaluation.} To visualize the degradation of truth directions across tasks of increasing difficulty, we project activations onto the 2D subspace spanned by the learned probe direction, and the direction of maximum variance in the orthogonal space (Figure~\ref{fig:truth_difficulty_scatter}; see Appendix~\ref{app:2d_projections} for more layers). The projection captures on average, across tasks and the two dimensions, $34.39\%$ of the total variance (see Appendix~\ref{app:2d_projection_and_var} for the projection methodology; Table~\ref{tab:variance_captured} for the detailed variance values).
From the arithmetic tasks, only A1 (at layer 25) shows separability of correct from incorrect expressions; activations in A2 and A3 seem increasingly intermixed.
For factual retrieval tasks (F0--F2), true and false statements form clearly separable clusters. However, \emph{as task complexity increases, the activations of true/false statements get more entangled.} This is visible from F3, 
while activations of tasks F4--F5 are almost indistinguishable. Since we operate in a high-dimensional space (here, 4096 dimensions), projecting activations in two dimensions only enables us to view the data from a limited viewpoint. By definition though, this angle is the one that best separates correct from incorrect statements, yet activations seem non-separable for the more difficult tasks (A2--A3 and F4--F5).
These results provide a finer-grained explanation for the generalization gap of truth directions observed in~\citet{bao-etal-2025-probing}, suggesting task complexity as a limiting factor for their generalization ability.

\section{Conclusion}
In this work, we systematically investigate the limits of linear truth directions in LLMs by varying the probing layer, task type and difficulty, and including model instructions for assessing correctness. We show that factual truth emerges early but disentangles from confounding features (e.g., sentence polarity) only in mid layers of the model, whereas arithmetic truth emerges continuously through late layers. Introducing explicit instructions for truth assessment in the prompts shifts the geometry of the learned truth probes and affects cross-task generalization. We identify a clear limitation of truth directions when truth assessment requires tracking intermediate results: the need for counting over a list of more than two elements within a known factual domain can break generalization to random chance. Our findings indicate that universality claims for truth directions are more limited than previously assumed; several choices can affect the geometry of the learned probe directions including the task type, model instructions, and layer of the model, while the difficulty of the task is a clear limiting factor. 

\section*{Limitations}

Our work has a number of limitations. Although we can find linear truth directions in most tasks, we probe only the final prompt token per layer, and we cannot rule out that truth is encoded nonlinearly or in a distributed manner within the network. In this paper, we attempt to isolate the effect of difficulty in our tasks. However, we cannot rule out the effect of surface features including statement length which grows (by construction of our data) with difficulty. Also, since features in language models may be encoded in superposition~\cite{elhage2022toymodelssuperposition}, our probes may converge to task-specific features, or in general, to features correlated with the truth label. Our analysis in Section~\ref{sec:layer_generalization} demonstrates one such case with the polarity of the statement. Further, we do not analyze model output behavior; relating truth directions to behavioral performance of the model is left to future work. Finally, we only tested models spanning 2B to 9B from the Llama and Gemma families, so it is unclear whether the same claims reproduce in larger models, different families, and in reasoning models.


\section*{Acknowledgments}

This research was funded by NSF award CNS-2319369. We thank Lucas Miguel Tassis, Aaron Mueller, and our research group at Boston University for helpful discussions throughout this project. We acknowledge the use of Claude Code for assistance with developing the experiments. The computational work reported on in this paper was performed on the Shared Computing Cluster which is administered by Boston University’s Research Computing Services.






\bibliography{references}
\appendix
\section{Experimental details}

\subsection{Dataset creation}\label{app:dataset_creation}
\paragraph{Factual knowledge datasets.}

The datasets \texttt{cities}, \texttt{neg\_cities}, and \texttt{cities\_conj} are adopted from \citet{marks_tegmark_2024}, containing unambiguous, curated true/false statements. We define six tasks of increasing intuitive difficulty as follows:
\begin{itemize}
    \item \textbf{F0 (cities):} Factual statements of the form ``The city of Paris is in France.''
    \item \textbf{F1 (neg\_cities):} Negated variants of F0, e.g., ``The city of Paris is \textit{not} in Australia.''
    \item \textbf{F2 (cities\_conj):} Conjunctions of two independent factual statements, e.g., ``It is the case both that the city of Paris is in France and the city of Dortmund is in Germany.''
    \item \textbf{F3 (cities\_same\_country\_quant):} Counting constraints over a list of 2 cities, e.g., ``Exactly 2 of the following cities are in France: Paris, Marseille.'' 
    \item \textbf{F4 (cities\_exact\_k):} Extends F3 by using a list of $N{=}5$ cities, e.g., ``Exactly 2 of the following cities are in Germany: Athens, Paris, Munich, Beijing, Dortmund.'' The true count $m$ of cities belonging to a randomly chosen country is computed; true examples state the correct $m$, false examples state $k \neq m$. Balanced across values of $k$.
    \item \textbf{F5 (cities\_exact\_k1\_k2):} Extends F4 to two countries simultaneously over a list of 6 cities, e.g., ``Exactly 2 of the following cities are in Germany and 1 in Greece: \dots''. True examples state the correct counts $(m_1, m_2)$; false examples use $(k_1, k_2) \neq (m_1, m_2)$.
\end{itemize}

\paragraph{Arithmetic datasets.}
Three arithmetic datasets of increasing task complexity are generated. All operands are integers sampled uniformly from $(0, 100)$. The four arithmetic operators used are $+, -, \times, /$; division is restricted to integer results. False examples present results that are perturbations of the correct result by a random offset in $[-10, -1] \cup [1, 10]$, so that an incorrect answer is not obvious.
\begin{itemize}
    \item \textbf{A1 (arith\_1op):} Single-operation expressions, e.g., $37 + 15 = 52$.
    \item \textbf{A2 (arith\_2ops):} Two-operation expressions, e.g., $(37 + 15) - 4 = 48$.
    \item \textbf{A3 (arith\_3ops):} Three-operation expressions, e.g., $(37 + 15) / (12 - 8) = 13$.
\end{itemize}

Table~\ref{tab:dataset_sizes} summarizes the number of examples per dataset. All datasets are balanced with respect to the class labels.

\begin{table}[ht]
\centering
\tiny
\begin{tabular}{llcc}
\toprule
\textbf{Task} & \textbf{Dataset name} & \textbf{Total examples} & \textbf{Train / Test} \\
\midrule
F0 & cities & 1{,}594 & 1{,}116 / 478 \\
F1 & neg\_cities & 1{,}594 & 1{,}116 / 478 \\
F2 & cities\_conj & 1{,}594 & 1{,}116 / 478 \\
F3 & cities\_same\_country\_quant & 2{,}000 & 1{,}400 / 600 \\
F4 & cities\_exact\_k & 2{,}000 & 1{,}400 / 600 \\
F5 & cities\_exact\_k1\_k2 & 2{,}000 & 1{,}400 / 600 \\
\midrule
A1 & arith\_1op & 1{,}000 & 700 / 300 \\
A2 & arith\_2ops & 1{,}000 & 700 / 300 \\
A3 & arith\_3ops & 1{,}000 & 700 / 300 \\
\bottomrule
\end{tabular}
\caption{Dataset sizes and train/test splits. All datasets are balanced with equal numbers of true and false examples.}
\label{tab:dataset_sizes}
\end{table}

\subsection{Model details}\label{app:sec:model_details}
We test whether our findings reproduce across models of different size and family. We select models of two different scales from  Llama~\citep{grattafiori2024llama3herdmodels} and Gemma~\citep{gemma2}. We use the instruction-tuned versions instead of the base models since they are more likely to follow instructions for truth evaluation when prompted. The main paper focuses on Llama-3.1-8B-Instruct; we provide results for the remaining three models in Appendix~\ref{app:sec:other_model_results}. Details of these models are shown in Table~\ref{app:tab_models}.

\begin{table}[ht]
\small 
\centering
\begin{tabular}{lccc}
\toprule
Model & $n_{\text{layers}}$ & $n_{\text{parameters}}$ & $d_{\text{model}}$ \\
\midrule
Llama-3.2-3B-Instruct  & 28 & 3.2B & 3072 \\
Llama-3.1-8B-Instruct  & 32 & 7.8B & 4096 \\
Gemma-2-2b-it          & 26 & 2.1B & 2304 \\
Gemma-2-9b-it          & 42 & 8.9B & 3584 \\
\bottomrule
\end{tabular}
\caption{Model specifications used in our experiments.}
\label{app:tab_models}
\end{table}

\subsection{Prompt templates}\label{app:prompts}

We define several instruction prompt templates to evaluate \textit{how truth directions in LLMs are affected by explicit instructions to assess correctness.}
The baseline is using no instruction as prefix to the prompted statement, i.e., the no-prompt setting which is common in prior work. We define the following variations, including generic question prompts such as \texttt{ask-correct}, and factual or arithmetic specific ones such as \texttt{ask-t/f}, \texttt{ask-able}, and \texttt{ask-arith} respectively. We have two factual-specific instruction prompts: \texttt{ask-t/f} asking for a 'True/False' answer and \texttt{ask-able} asking for a 'Yes/No' answer. This allows us to test whether the answer we ask to be generated is affecting how the model operates to decide correctness of the input statement and hence affects truth directions.
\begin{tcolorbox}[
    colback=Maroon!7,
    colframe=Maroon,
    colbacktitle=Maroon,
    coltitle=white,
    fonttitle=\bfseries,
    title=Prompt templates for model instructions,
    rounded corners,
    boxrule=1pt,
    left=6pt, right=6pt, top=4pt, bottom=4pt
]
\textbf{no-prompt:}~\texttt{The city of Paris is in France.}\\[4pt]
\textbf{ask-correct:}~\texttt{Is the following correct?\,${\scriptstyle\hookleftarrow}$\,The city of Paris is in France. Answer:}\\[4pt]
\textbf{ask-t/f:}~\texttt{Is the following statement TRUE or FALSE?\,${\scriptstyle\hookleftarrow}$\,The city of Paris is in France. Answer:}\\[4pt]
\textbf{ask-able:}~\texttt{Are you able to evaluate the following statement according to its truthfulness?\,${\scriptstyle\hookleftarrow}$\,The city of Paris is in France. Answer:}\\[4pt]
\textbf{ask-arith:}~\texttt{Are you able to evaluate the following arithmetic expression according to its correctness?\,${\scriptstyle\hookleftarrow}$\,13+71=84. Answer:}
\end{tcolorbox}

\subsection{Probe training details}\label{app:sec:probe_training}
We train logistic regression probes on activations extracted from the final token at a given layer. We mean-center activations and train a no-bias probe so that we get a direction passing through the origin. We train using the Adam optimizer~\citep{kingma2017adammethodstochasticoptimization} with learning rate $10^{-3}$ and weight decay $0.1$ to minimize the binary cross-entropy loss over 1000 training steps. We split each dataset into $70\%$ train and $30\%$ test sets.

\subsection{Projection methodology and captured variance}~\label{app:2d_projection_and_var}

For a given task and layer $\ell$, let $h_i^{(\ell)}\in\mathbb{R}^{d_\text{model}}$ be the residual-stream activation at the final token of example $i$, and let $\hat{w}^{(\ell)}$ be the unit-normalized truth direction and $\mu^{(\ell)}$ the mean of its training set. We mean-center the activations, $\tilde{h}_i^{(\ell)} = h_i^{(\ell)} - \mu^{(\ell)}$.

The horizontal coordinate is the projection onto the truth direction $\hat{w}^{(\ell)}$,
\[
a_i^{(\ell)} = \tilde{h}_i^{(\ell)} \hat{w}^{(\ell)},
\]
We subtract this component to obtain the part of the activation orthogonal to it,
\[
r_i^{(\ell)} = \tilde{h}_i^{(\ell)} - a_i^{(\ell)}\, \hat{w}^{(\ell)} .
\]

The vertical axis is the direction of maximum variance in this orthogonal space, which we find as follows.
Let $R^{(\ell)}\in\mathbb{R}^{n\times d_\text{model}}$, be the matrix of the residual activations $r_i^{(\ell)}$ for all examples $i=1,\dots,n$ at layer $\ell$. We take $v^{(\ell)}$ to be the top eigenvector of $R^{(\ell)\top}R^{(\ell)}$. Then,
\[
b_i^{(\ell)} = r_i^{(\ell)} v^{(\ell)}.
\]

is the coordinate for the vertical axis of the projection.

Each activation is shown as the pair $\big(a_i^{(\ell)},b_i^{(\ell)}\big)$: its position
along the learned truth direction and along the axis of largest variation in the orthogonal space. 

\begin{table}
    \centering
    \small
    \caption{Fraction of variance in model activations (layer 25) captured by the 2D projection subspace used in Figure~\ref{fig:truth_difficulty_scatter}, spanned by the truth probe direction and the direction of maximum variance orthogonal to it.}
    \label{tab:variance_captured}
    \begin{tabular}{lrrr}
        \toprule
        \textbf{Task} & \textbf{Truth dir.} & \textbf{Max-var dir.} & \textbf{Cumulative} \\
        \midrule
        A1 &  1.01\% & 32.56\% & 33.57\% \\
        A2 &  0.67\% & 35.95\% & 36.62\% \\
        A3 &  0.49\% & 28.17\% & 28.66\% \\
        F0 & 17.01\% & 29.25\% & 46.26\% \\
        F1 & 13.62\% &  8.51\% & 22.13\% \\
        F2 & 29.30\% & 18.46\% & 47.76\% \\
        F3 &  3.28\% & 11.63\% & 14.92\% \\
        F4 &  1.12\% & 42.63\% & 43.74\% \\
        F5 &  1.06\% & 34.75\% & 35.82\% \\
        \midrule
        \textit{Mean} & \textit{7.51\%} & \textit{26.88\%} & \textit{34.39\%} \\
        \bottomrule
    \end{tabular}
\end{table}

\section{Supplementary results}

\subsection{Layer selection with variance ratio}
\label{app:layer-selection}
\begin{figure}[ht]
    \centering
    \includegraphics[width=\linewidth]{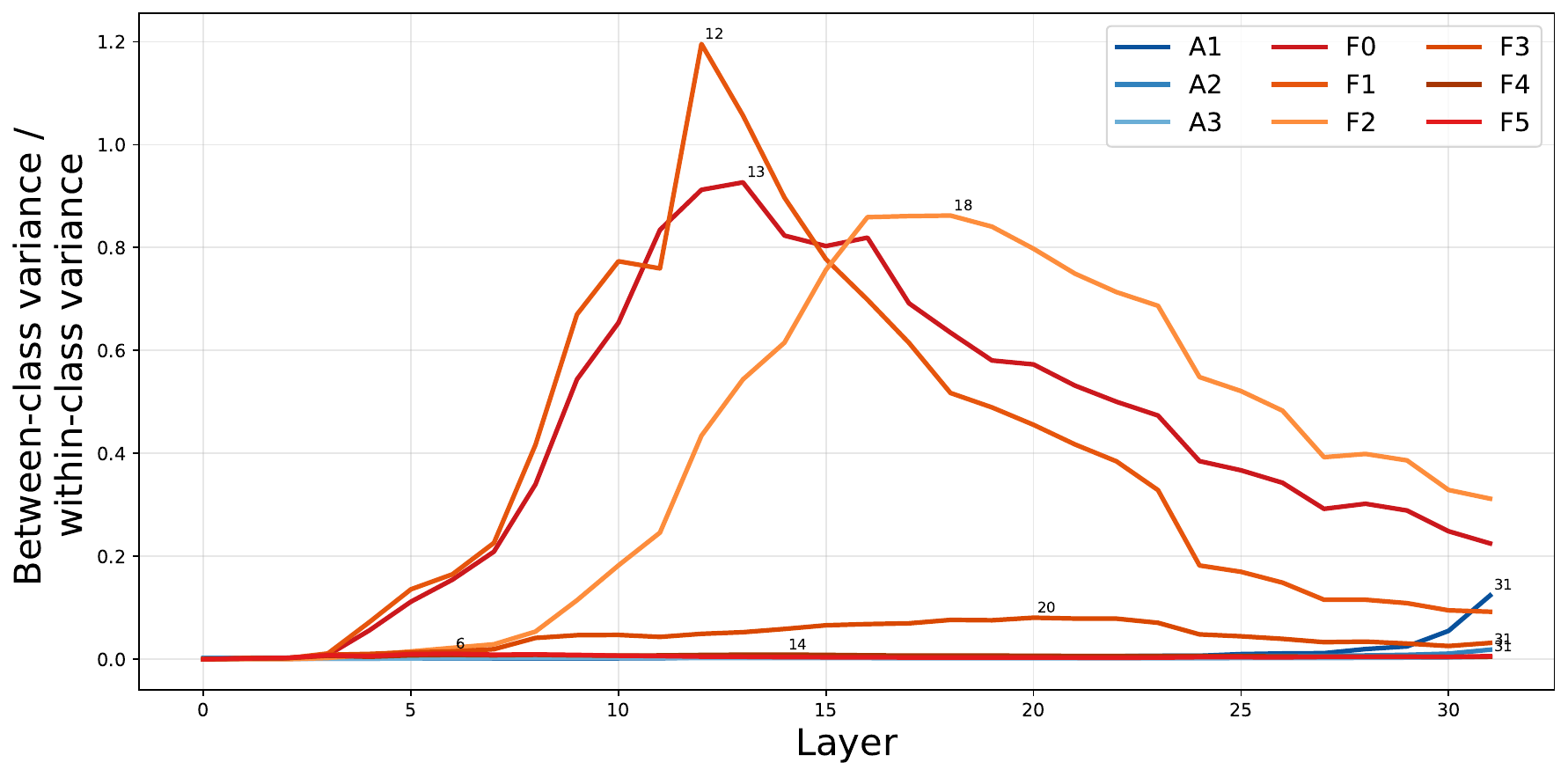}
    \caption{Between- to within-class variance ratio $R^{(\ell)}$ across layers for each task.}
    \label{fig:app_variance_ratio}
\end{figure}

Prior work~\citep{burger_2024, bao-etal-2025-probing} selects the layer for probing activations by maximizing the ratio of between to within class variance of activations. For layer $\ell$ with class means $\mu_{\text{true}}^{(\ell)}, \mu_{\text{false}}^{(\ell)}$, overall mean $\mu^{(\ell)}$, this ratio is defined as follows:
\[
R^{(\ell)} = \frac{\|\mu_{\text{true}}^{(\ell)} - \mu^{(\ell)}\|^2 + \|\mu_{\text{false}}^{(\ell)} - \mu^{(\ell)}\|^2}{\text{Var}(h_i^{(\ell)} : y_i = 1) + \text{Var}(h_i^{(\ell)} : y_i = 0)}
\]
where $\|\cdot\|^2$ is the mean squared difference across dimensions. We compute this ratio for our datasets across layers. Figure~\ref{fig:app_variance_ratio} shows that the ratio peaks at different layers for different tasks, confirming that no single layer is universally optimal for extracting truth directions.

\subsection{Polarity truth direction over layers}\label{app:polarity}

\begin{figure}[ht]
    \centering
    \includegraphics[width=\linewidth]{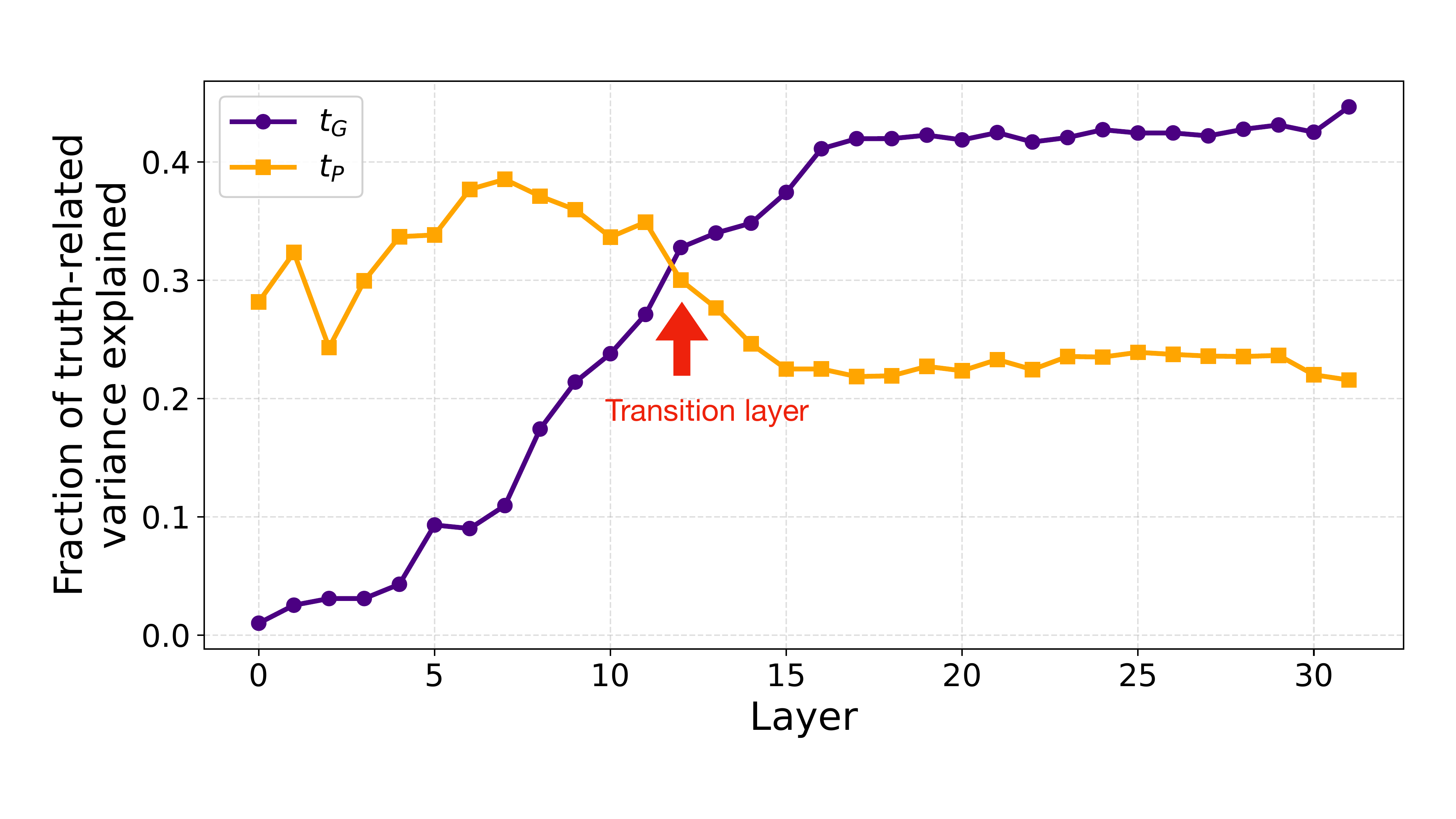}
    \caption{Fraction of truth-related variance explained by the polarity-invariant direction $t_G$ and the polarity-dependent direction $t_P$ across layers.}
    \label{fig:app_t_g_t_p_variance}
\end{figure}

Figure~\ref{fig:app_t_g_t_p_variance} shows the fraction of truth-related variance explained by the directions $t_G$ and $t_P$ as analyzed in \citet{burger_2024}. At layer 12 (the layer analyzed in their work), both directions explain similar fractions ($\sim$0.33 each), consistent with their two-dimensional subspace finding. In early layers, $t_P$ dominates, confirming that early-layer probes primarily capture sentence polarity. By layer 16, $t_G$ explains more variance and at the same time $t_P$ decays, suggesting that the two-dimensional subspace is happening in this transitional phase of model computation and is not a universal property of truth directions in LLMs.

\subsection{Results across other prompt templates}\label{app:sec:other_prompts_results}

In this subsection we report results for all the prompt templates defined in Appendix~\ref{app:prompts}.

\subsubsection{AUROC per layer across prompt templates}
\begin{figure}[ht]
    \centering
    \includegraphics[width=\linewidth]{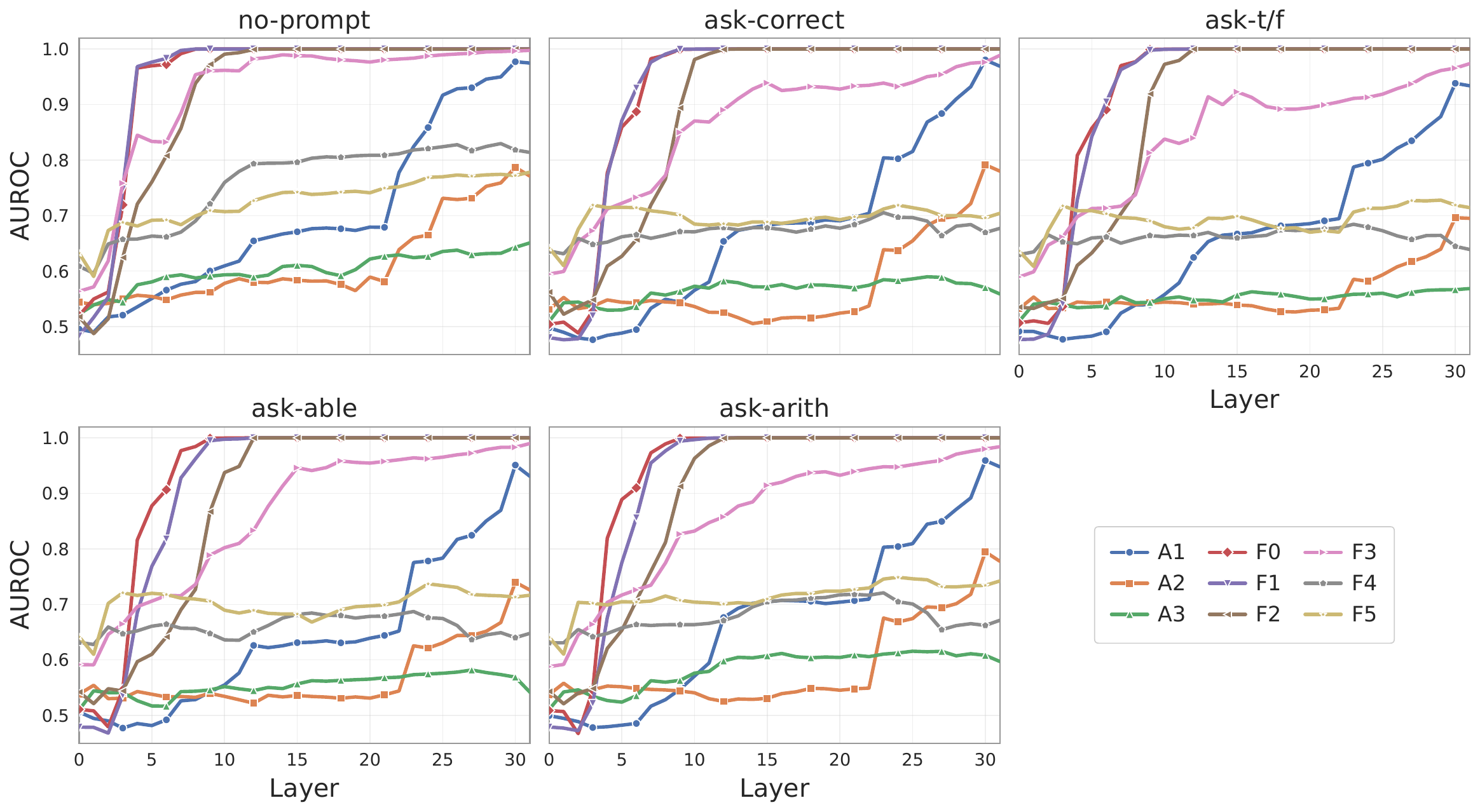}
    \caption{In-domain test-set AUROC per layer across all prompt templates.}
    \label{app:auroc_per_layer_prompts}
\end{figure}

Figure~\ref{app:auroc_per_layer_prompts} shows the in-domain test-set $\metric$ per layer for the five prompt templates we test. Although truth directions emerge in all prompts, model instructions affect truth direction emergence patterns. Specifically, in some cases they delay the emergence of truth directions (strongly for F3 across prompt templates) but also might result in lower peak performance compared to the no-prompt setting.

\subsubsection{Cosine similarity of truth probes across prompt templates} 
\begin{figure*}[h]
    \centering
    \includegraphics[width=.9\linewidth]{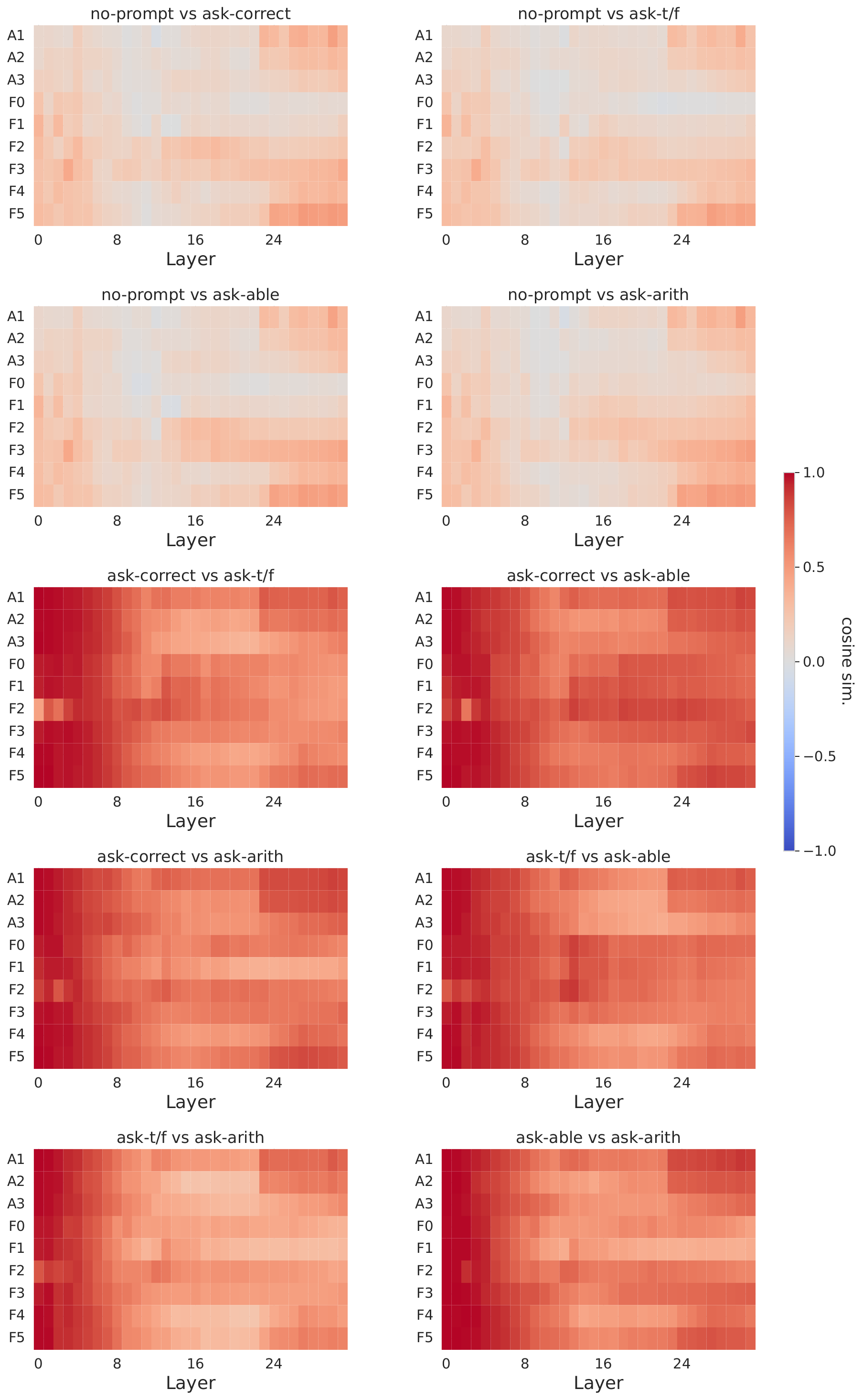}
    \caption{Cosine similarity between truth probes trained under different prompt templates across layers and tasks for Llama-3.1-8B-Instruct.}
    \label{fig:app_cosine_all_pairs}
\end{figure*}

Figure~\ref{fig:app_cosine_all_pairs} shows the cosine similarity between truth directions trained using all pairs of combinations of the five prompts. The first two rows of heatmaps show the cosine similarity of pairs between no-prompt and all explicit instruction prompts, showing mostly low cosine similarity. On the other hand, any pair of explicit instruction prompt probes shows high cosine similarity despite different wording of the instructions.

\subsubsection{Cross-task generalization of truth probes across prompt templates}

\begin{figure*}[h]
    \centering
    \includegraphics[width=\textwidth]{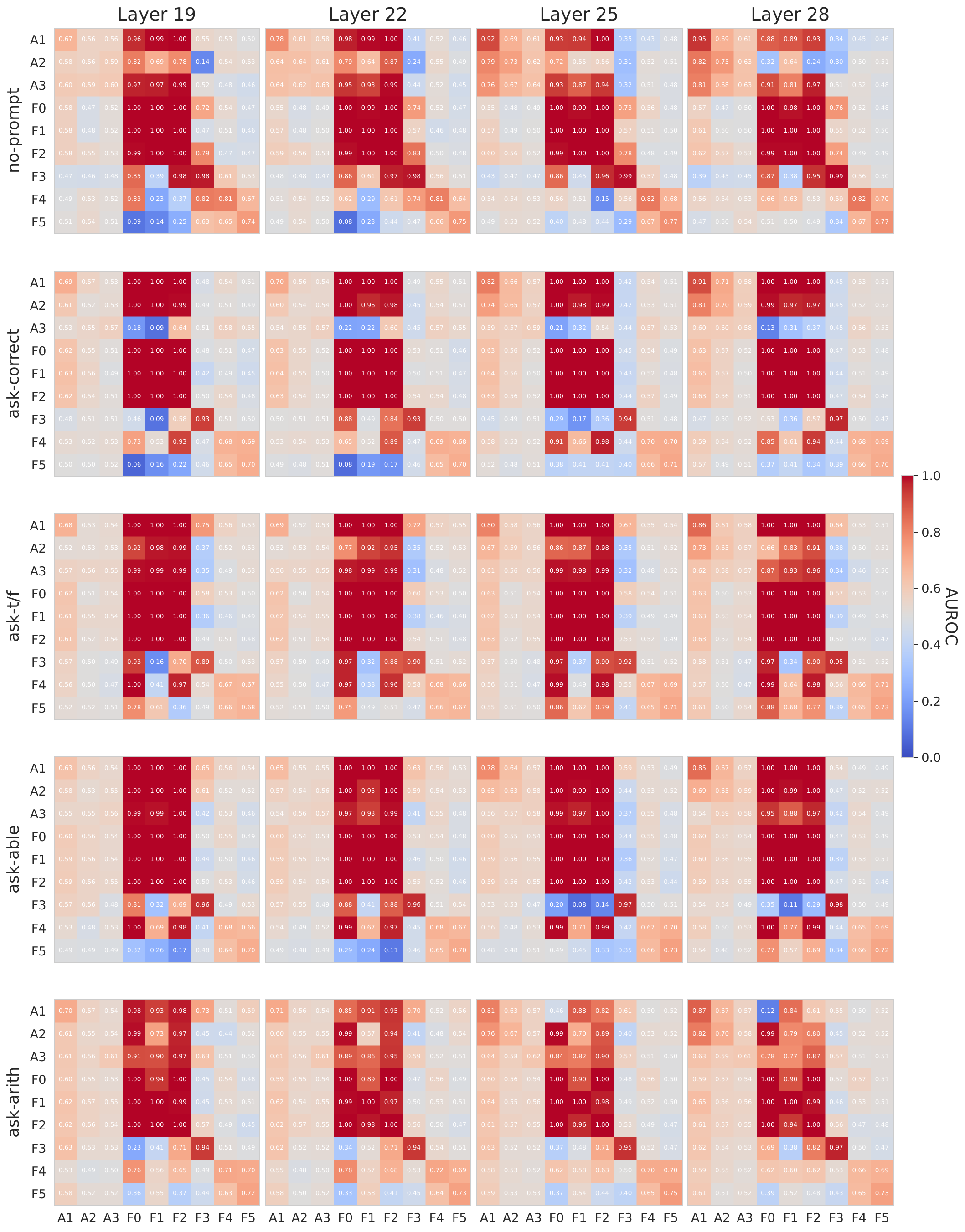}
    \caption{Cross-task generalization for all five prompts at layer depth fractions 0.6, 0.7, 0.8 and 0.9 for \textbf{Llama-3.1-8B-Instruct}. Rows indicate training task, columns indicate test task.}
    \label{app:cross_task_generalization}
\end{figure*}

Figure~\ref{app:cross_task_generalization} shows the generalization heatmaps across all five prompt settings at a sample of layers. The patterns we analyzed in the main paper hold across all templates: pure factual to factual generalization (F0--F2) is consistently high, arithmetic-trained probes show in most cases higher generalization to F0--F2 under instruction prompts, and generalization to the harder factual tasks (F3--F5) remains low regardless of the prompt template used. The ask-arith template shows weaker generalization to F0--F2 compared to the other instruction prompts; the specialized arithmetic instruction does not seem to produce the same effect as a more general truth assessment instruction.

\subsection{Tasks of intermediate difficulty}\label{app:intermediate_difficulty}

\begin{figure*}[t]
  \centering
  \includegraphics[width=.8\linewidth]{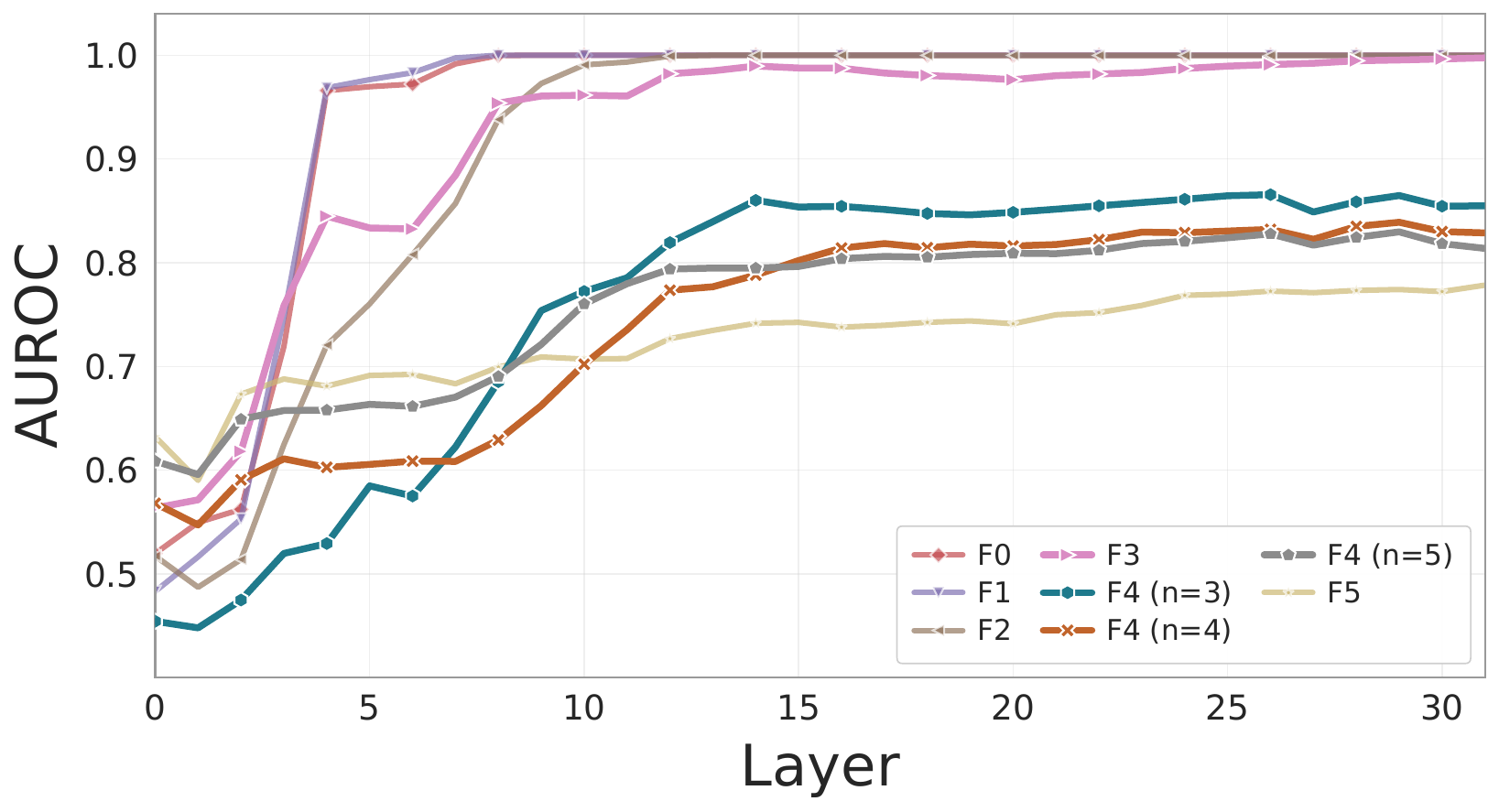}
  \caption{In-domain test AUROC per layer for factual tasks of intermediate difficulty. The tasks of intermediate difficulty with varying list lengths are shown with emphasized lines.}
  \label{fig:auroc-factual-intermediate}
\end{figure*}

Probes trained on F3 achieve $\metric\approx1.0$, whereas those trained on F4 drop to $\metric\approx0.8$. We want to understand \emph{what operation introduces this difficulty.} It might be the list length, or the need for counting itself. To investigate this, we create intermediate variants of the task F4 where we vary the list length from 3 cities, up to 5 (the original task). Note that, when the list length is 2 the model could exploit the shortcut of only checking whether both cities belong to the stated country, as tested in the task F2; the model successfully computes truth for F3, which indicates that either it counts up to two, or it exploits such a shortcut. Figure~\ref{fig:auroc-factual-intermediate} shows that \textbf{the performance drop happens as soon as we increase the list length to 3}, and increasing further to 4 and 5 we observe very little additional degradation. This suggests that the need for genuine counting how many (even, out of three) cities satisfy the condition introduces the limitation.

\subsection{Control experiments}

\subsubsection{Token control experiment}~\label{app:token_control}

Since the probing token differs across \texttt{no-prompt}, and \texttt{ask-correct} settings, we measure to what extent the obtained truth directions change when we change the last token of \texttt{ask-correct} to match the one in \texttt{no-prompt}. To do so, we change the final token of \texttt{ask-correct} to be a period token instead of a colon. Further, we test whether the ``Answer'' cue in the prompt has an effect on our results or is solely the instruction prefix (i.e., ``Is the following correct?'') that matters. The two control prompt templates are as follows.

\begin{tcolorbox}[
    colback=Maroon!7,
    colframe=Maroon,
    colbacktitle=Maroon,
    coltitle=white,
    fonttitle=\bfseries,
    title=ask-correct control templates,
    rounded corners,
    boxrule=1pt,
    left=6pt, right=6pt, top=4pt, bottom=4pt
]
\textbf{ask-correct-period:}~\texttt{Is the following correct?\,${\scriptstyle\hookleftarrow}$\,The city of Paris is in France. Answer.}\\[4pt]
\textbf{ask-correct-prefix:}~\texttt{Is the following correct?\,${\scriptstyle\hookleftarrow}$\,The city of Paris is in France.}
\end{tcolorbox}

\begin{figure*}[h]
    \centering
    \includegraphics[width=\linewidth]{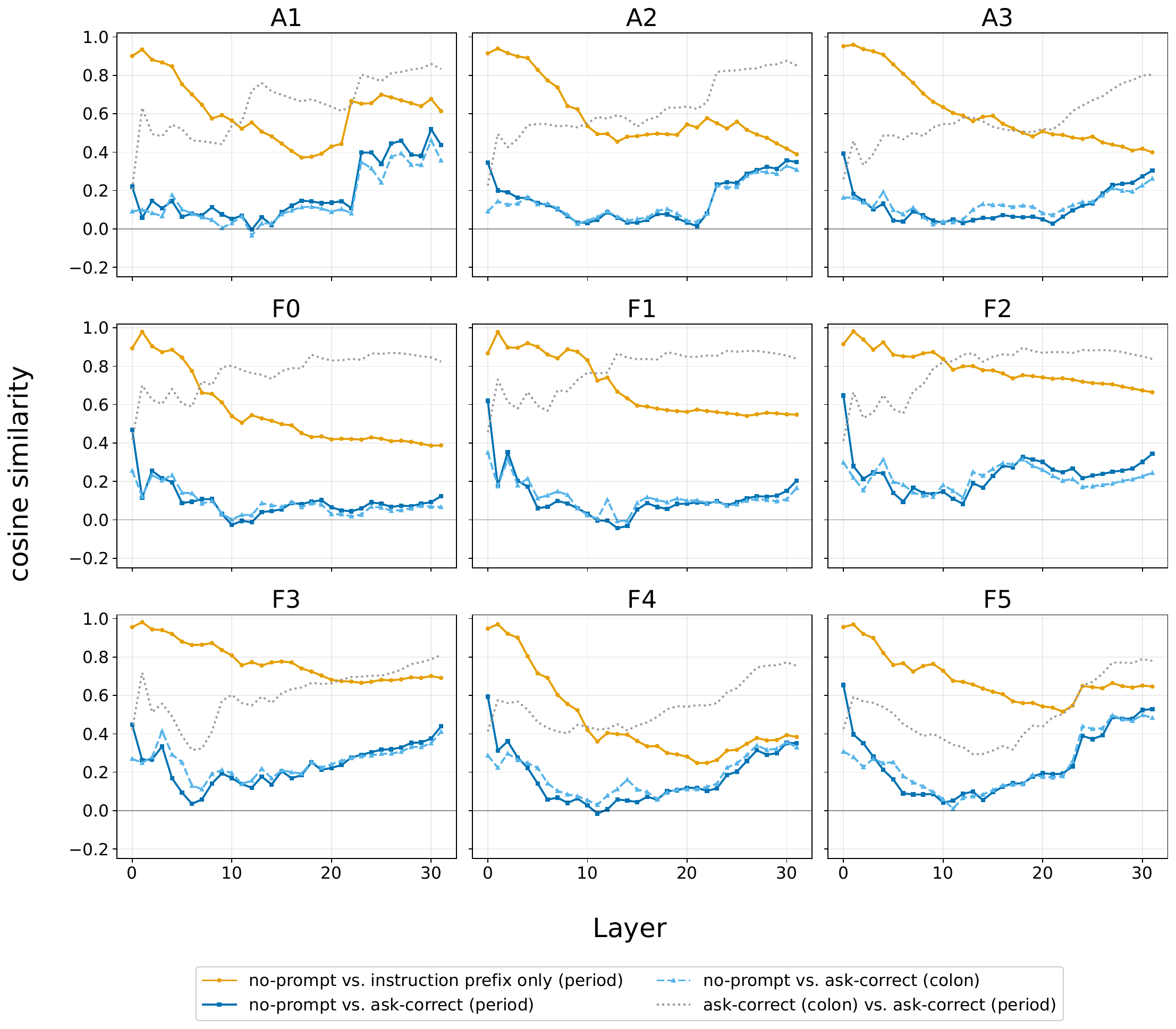}
    \caption{Cosine similarity of truth directions probed on different tokens of ask-correct instruction template compared with no-prompt probes.}
    \label{app:fig:token_control_cosines}
\end{figure*}

Figure~\ref{app:fig:token_control_cosines} shows the cosine similarity of \texttt{no-prompt} truth directions with the \texttt{ask-correct-prefix} in orange, \texttt{ask-correct-period} in solid blue, and the \texttt{ask-correct} in light dashed blue. Across tasks we observe the following. Truth directions trained under ask-correct with either a period or a colon as the last token show high cosine similarity with each other (dotted gray lines), especially in the later layers where truth directions emerge. This indicates that \textit{the identity of the last token has limited effect on the geometry of the learned truth directions in our templates.} Further, the two blue lines are extremely close to each other across all tasks and layers, indicating that the geometric effect of ask-correct instructions on truth directions compared to no-prompt, is not influenced by whether the last token is a period or a colon. Finally, we observe that omitting the ``Answer'' suffix yields truth directions that are highly aligned with the no-prompt setting, suggesting that the instruction prefix alone does not suffice to produce the effects we observe in the main paper.

\subsubsection{Prompt template control experiment}\label{app:control_experiment}

The shift in the geometry of the learned truth probes observed under explicit task framing raises the question of whether this effect is caused solely from the instruction for truth assessment, or by other variables such as the introduction of (possibly random) extra tokens or the instruction-following context of the prompt. To investigate this, we repeat our analysis with two additional prompt templates shown below, with same token length as the \texttt{ask-correct} prompt. 

\begin{figure*}[h]
    \centering
    \includegraphics[width=.95\linewidth]{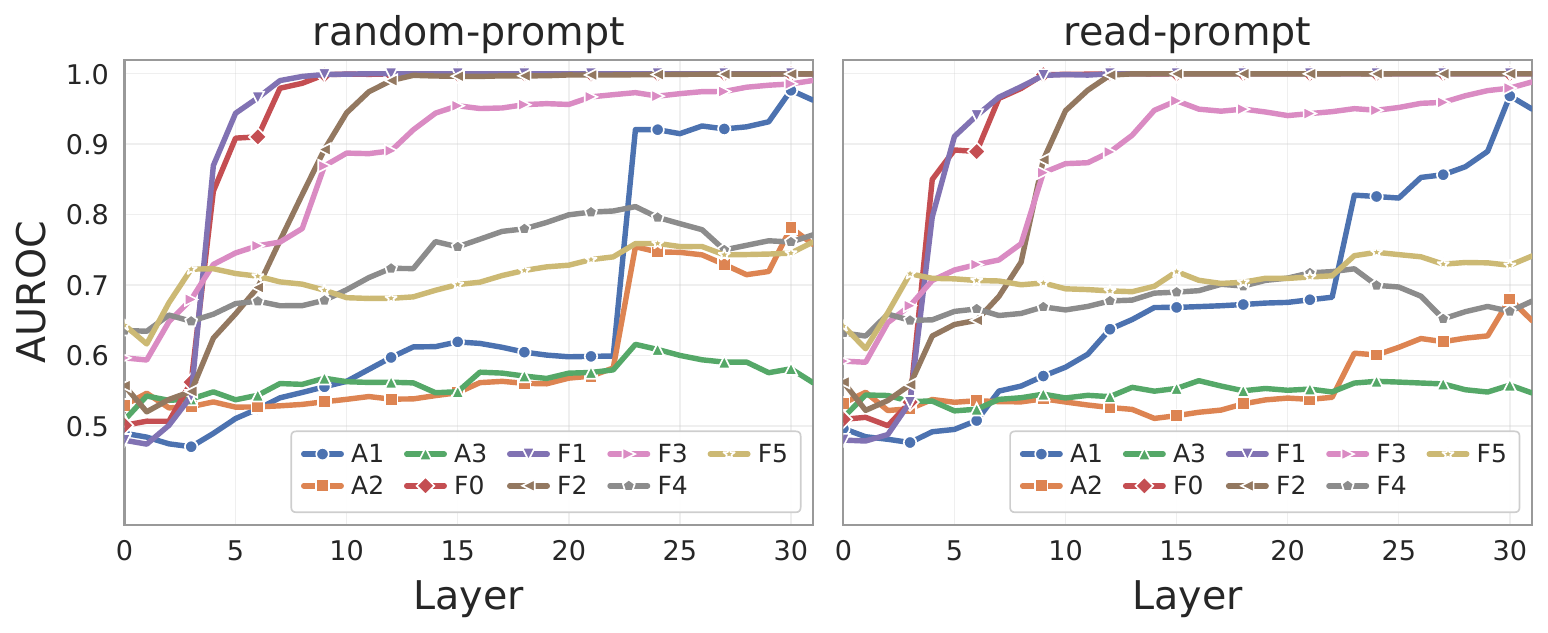}
    \caption{Test-set AUROC per layer for control prompts.}
    \label{app:fig:control_prompts_auroc}
\end{figure*}

\begin{figure*}[t]
    \centering
    \begin{subfigure}{0.48\linewidth}
        \centering
        \includegraphics[width=\linewidth]{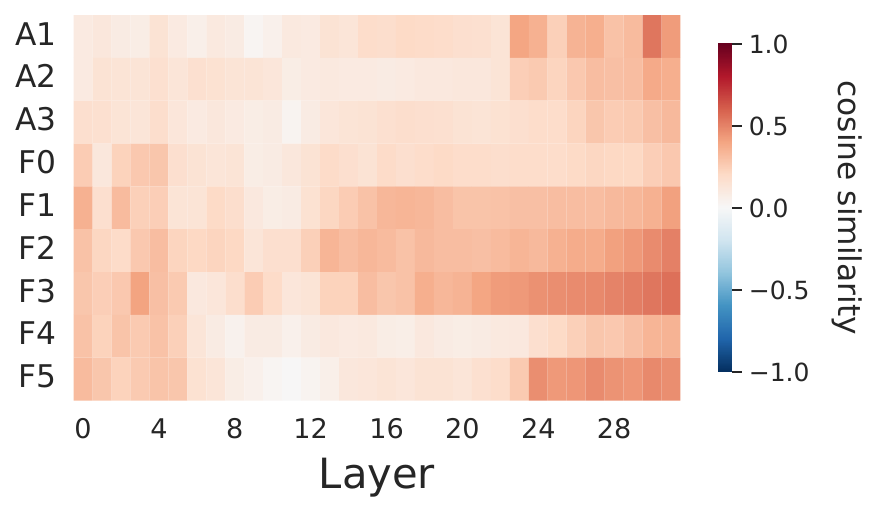}
        \caption{Cosine similarity of no-prompt probes with random-prompt probes across layers.}
        \label{app:fig:random_prompt_cosine}
    \end{subfigure}
    \hfill
    \begin{subfigure}{0.48\linewidth}
        \centering
        \includegraphics[width=\linewidth]{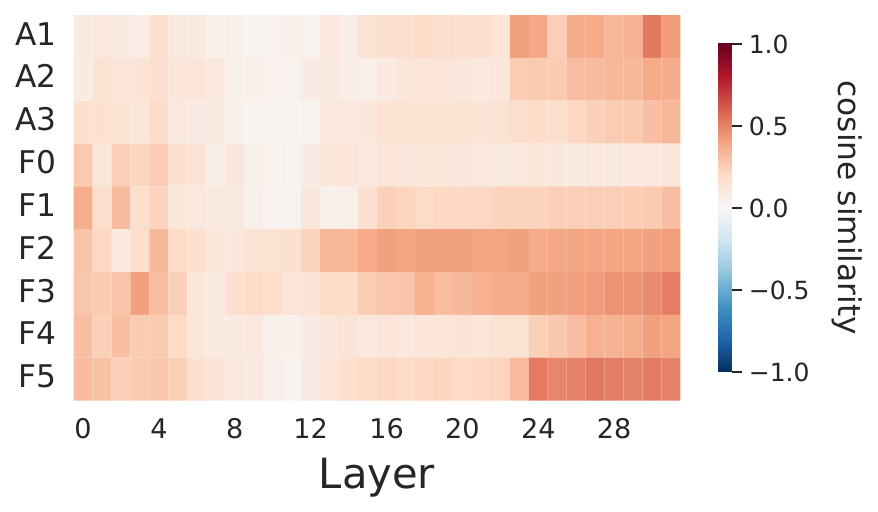}
        \caption{Cosine similarity of no-prompt probes with read-prompt probes across layers.}
        \label{app:fig:read_prompt_cosine}
    \end{subfigure}
    \caption{Cosine similarities of control probes with no-prompt probes across layers.}
    \label{app:fig:control_prompts_cosine}
\end{figure*}

\begin{figure*}[t]
    \centering
    \includegraphics[width=\linewidth]{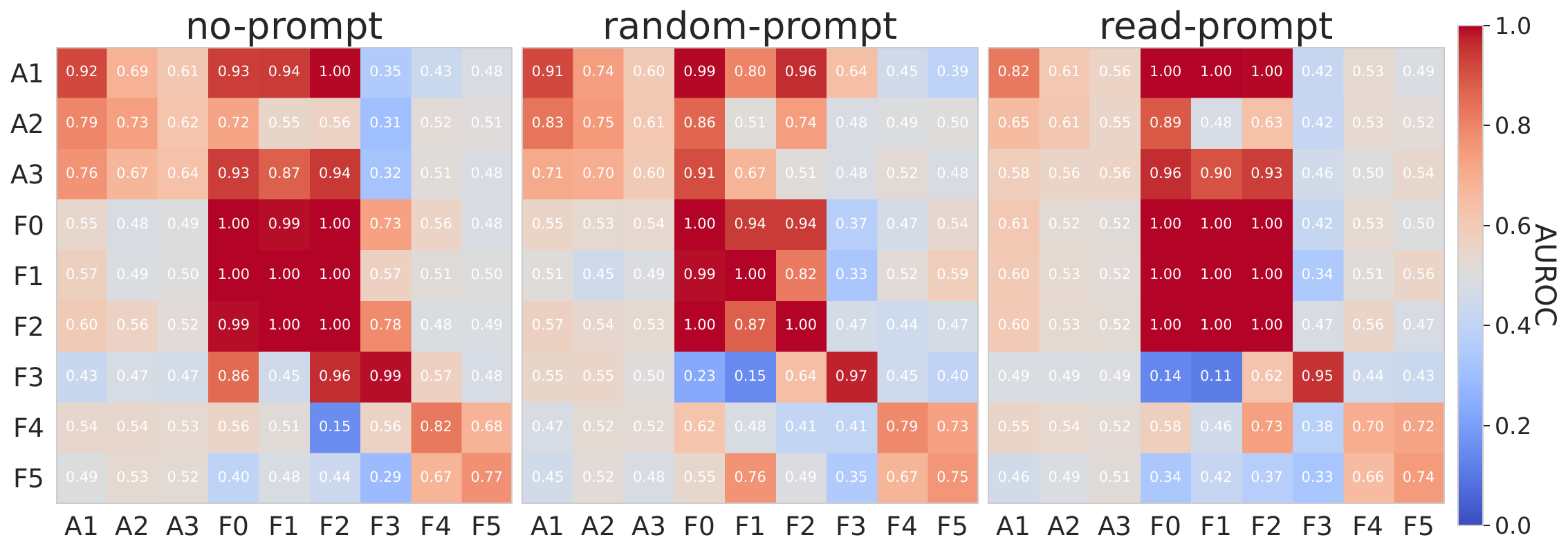}
    \caption{Cross-task generalization of probes trained with no-prompt and the control prompts.
    }
    \label{app:fig:control_prompts_generalization}
\end{figure*}

Figure~\ref{app:fig:control_prompts_auroc} shows that under both control prompts truth directions emerge consistently, similarly to what we observe across all prompt templates we test.
Figure~\ref{app:fig:control_prompts_cosine} compares each control prompt probe with no-prompt probe direction per layer to measure to what extent these templates affect the geometry of the learned truth probes. Both control templates yield directions with low cosine similarity to the no-prompt directions in most layers; however they show higher similarity compared to what we observe for truth assessment instructions in Figure~\ref{fig:app_cosine_all_pairs}. Therefore, the control templates seem to change --although less so, compared to truth assessment instructions-- the geometry of the learned truth probes. Finally, in Figure~\ref{app:fig:control_prompts_generalization} we test how cross-task generalization is affected. We observe that the effect looks similar to no-prompt probes. \emph{Although the introduction of these prefix tokens can cause a shift in the geometry of truth probes, they do not produce the same effect observed using instructions for truth assessment.}

\begin{tcolorbox}[
    colback=Maroon!7,
    colframe=Maroon,
    colbacktitle=Maroon,
    coltitle=white,
    fonttitle=\bfseries,
    title=Prompt control templates,
    rounded corners,
    boxrule=1pt,
    left=6pt, right=6pt, top=4pt, bottom=4pt
]
\textbf{random-prompt:}~\texttt{Green table running bright.\,${\scriptstyle\hookleftarrow}$\,The city of Paris is in France. Answer:}\\[4pt]
\textbf{read-prompt:}~\texttt{Read the following sentence.\,${\scriptstyle\hookleftarrow}$\,The city of Paris is in France. Answer:}
\end{tcolorbox}

\subsection{Additional activation projections}\label{app:2d_projections}
\begin{figure*}[h]
\centering
{\includegraphics[width=\linewidth]{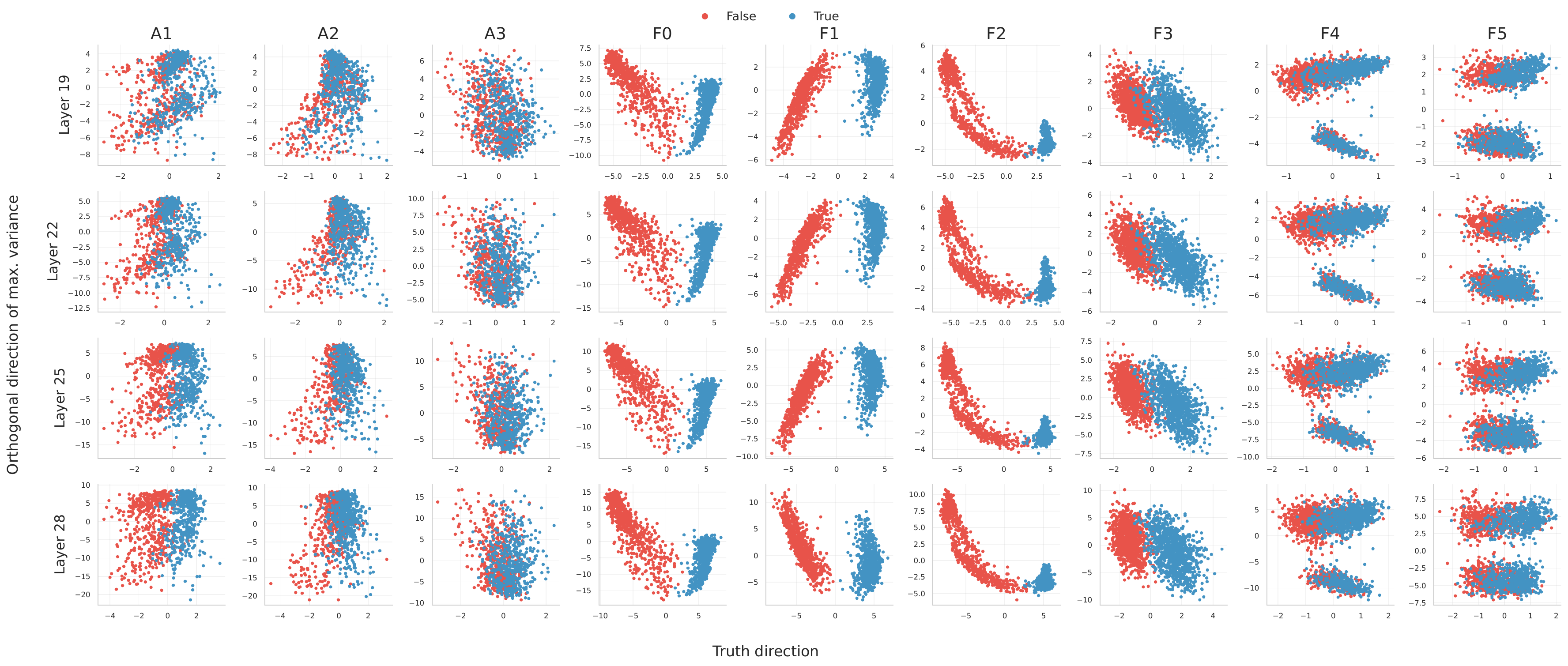}}

\caption{Activation projections for \textbf{Llama-3.1-8B-Instruct}.}
\label{app:fig:projections_llama8b}
\end{figure*}

In Figure~\ref{app:fig:projections_llama8b} we show supplementary projections of the activations over the two-dimensional subspace defined in Section~\ref{sec:difficulty}. We select a subset of model layers as fractions of the model depth: 0.6, 0.7, 0.8, 0.9.

\section{Results for other models}\label{app:sec:other_model_results}

Below we show results for the rest of the models we tested. The difficulty boundary for truth directions identified in the main paper replicates across all four models; generalization to harder factual tasks (F3--F5) remains consistently low regardless of model instructions, model size or model family. Also, within-family factual generalization (F0--F2) is consistently strong across all models and prompt settings. The prompt effect on cross-task generalization holds across both model families. However, the generalization improvement observed in Llama models is observed from A1--A3 $\rightarrow$ F0--F2; in Gemma models the effect is more pronounced on F3--F5 $\rightarrow$ F0--F2. This suggests that the particular effect caused by model instructions may vary across model families.

\subsection{Llama-3.2-3B-Instruct}

\begin{figure*}[h]
\centering
\begin{minipage}[t]{0.42\textwidth}
    \vspace{0pt}
    \subfloat[In-domain test AUROC across layers.]{\includegraphics[width=.95\linewidth]{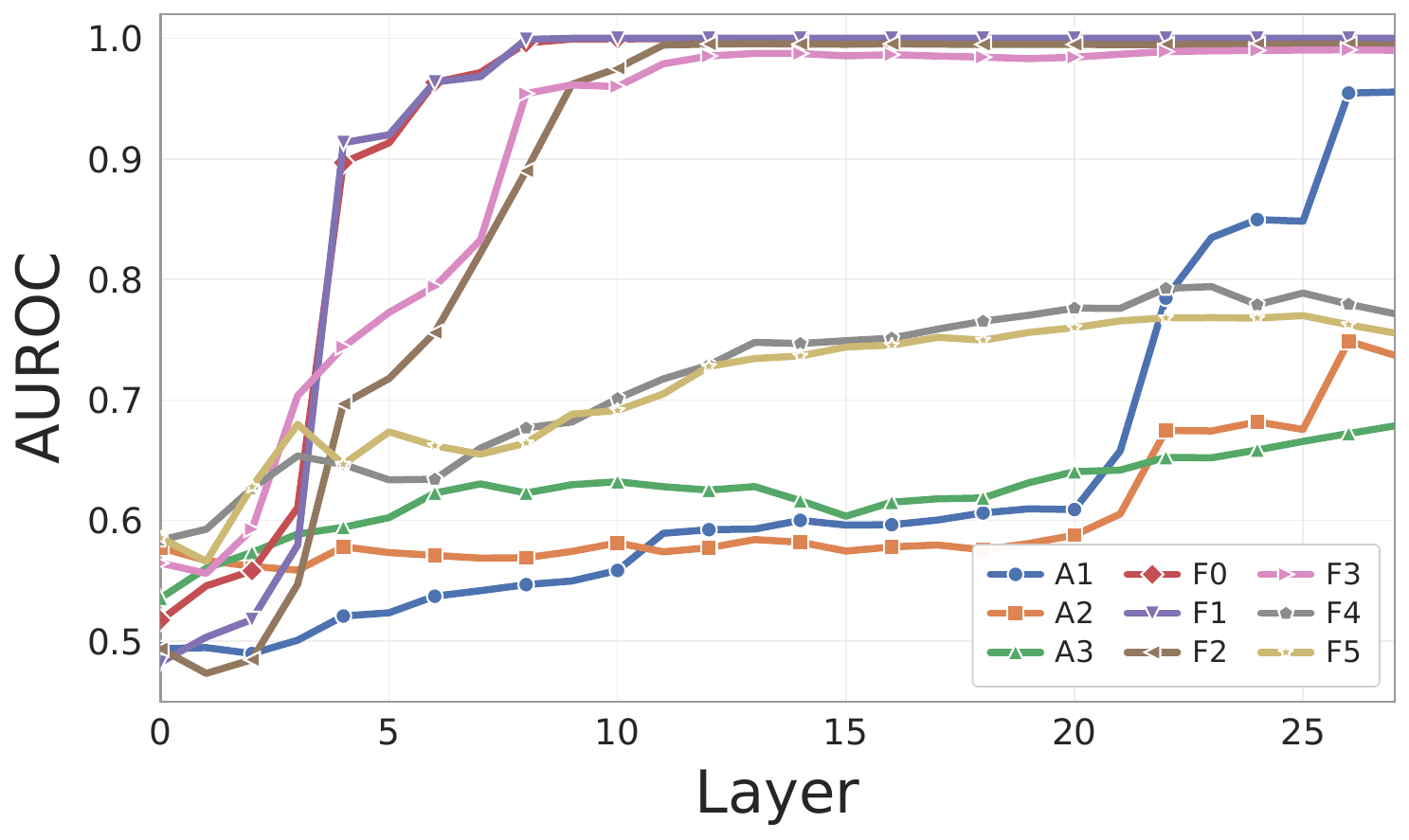}
    }\\[0.5em]
    \subfloat[F0-trained probe cross-task generalization across layers.]{\includegraphics[width=.95\linewidth]{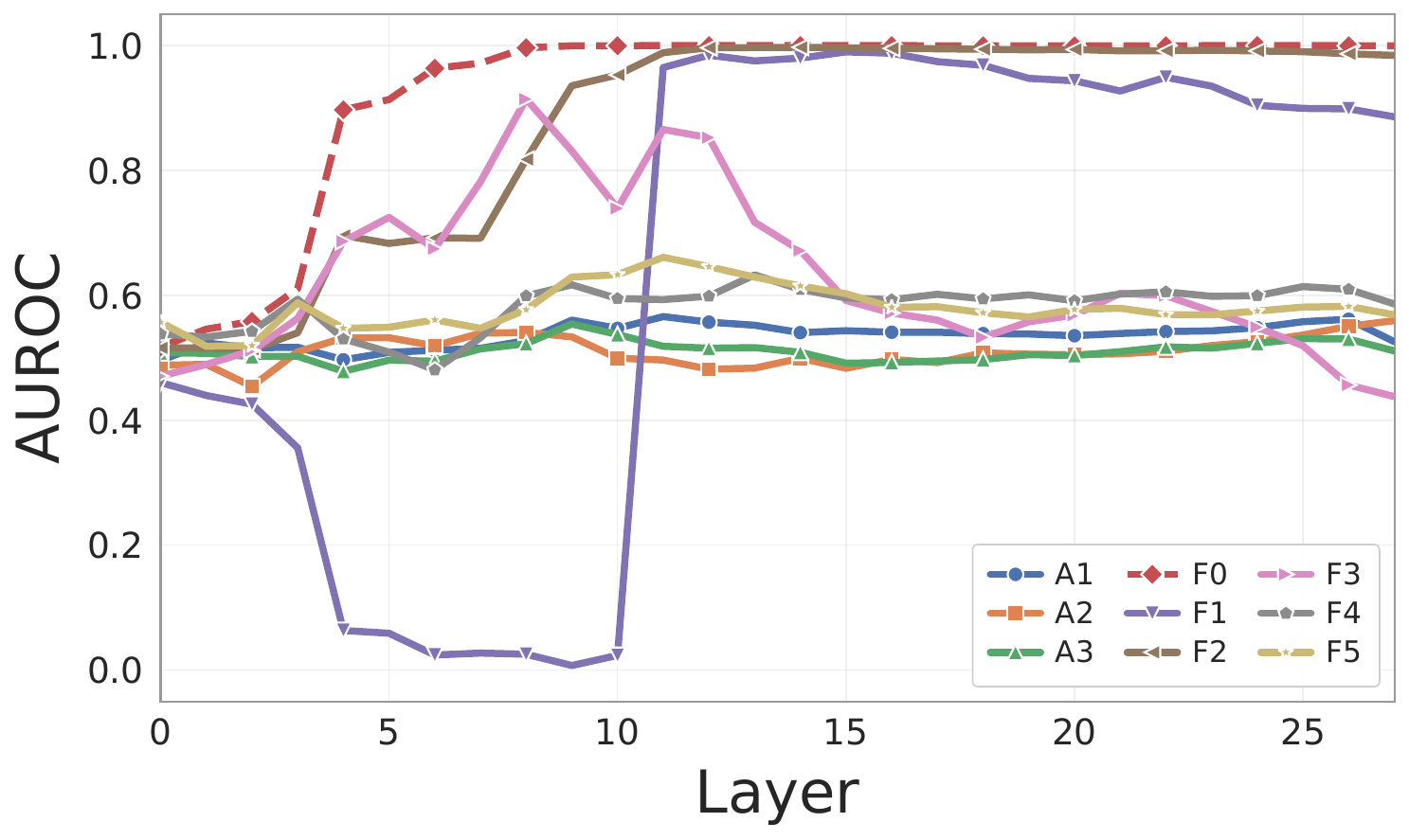}}
\end{minipage}
\hfill
\begin{minipage}[t]{0.56\textwidth}
    \vspace{0pt}
    \subfloat[Cosine similarity of probes across all pairs of layers for each task.]{%
        \includegraphics[width=\linewidth]{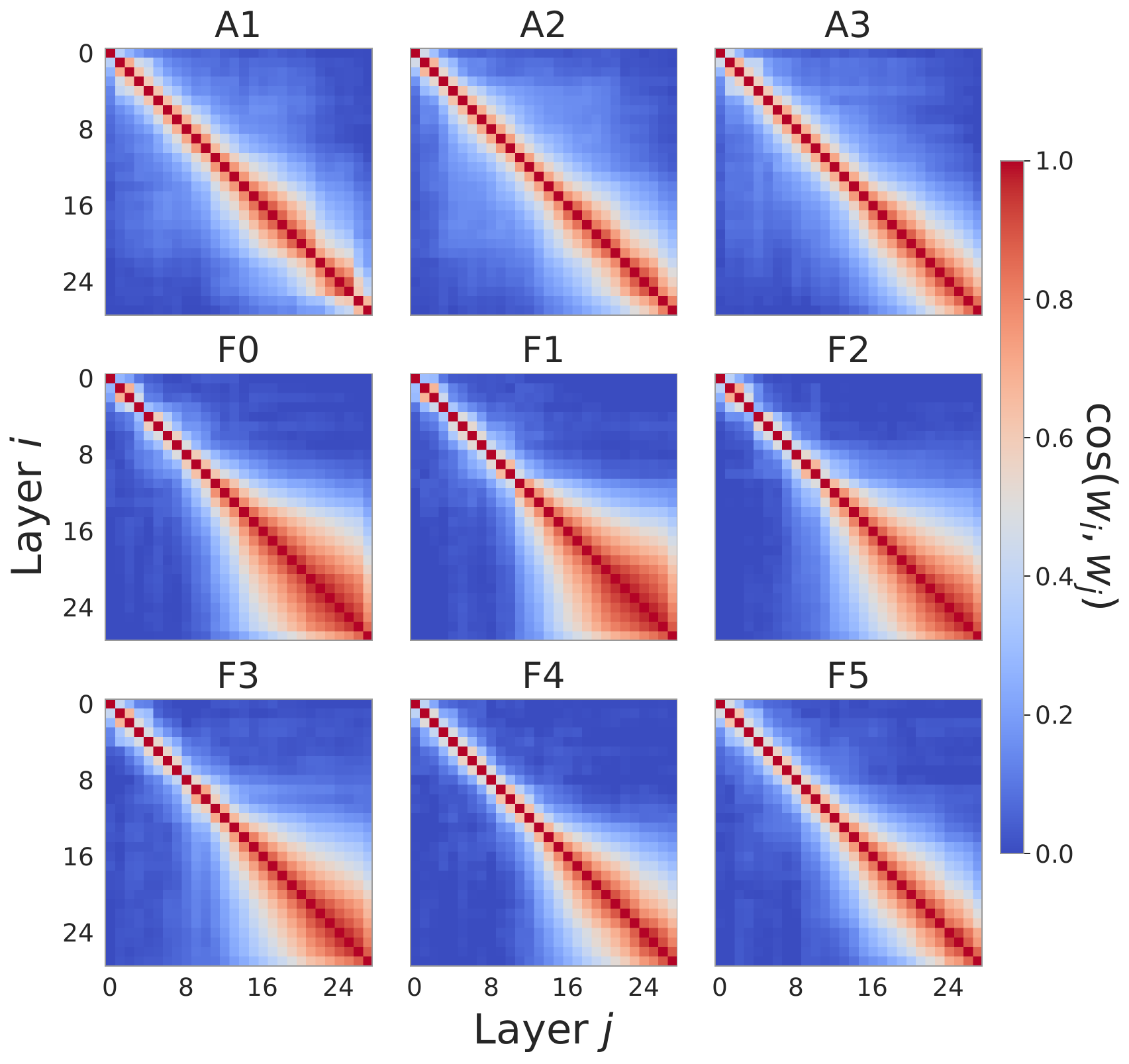}%
    }
\end{minipage}
\caption{Layer dependence of truth directions for \textbf{Llama-3.2-3B-Instruct}. (a) In-domain test-set performance. (b) F0-probe cross-task generalization. (c) Cosine similarity of probes across layer pairs.}
\label{app:fig:layer_dependence_combined_llama3b}
\end{figure*}

\begin{figure*}[h]
    \centering
    \includegraphics[width=.9\linewidth]{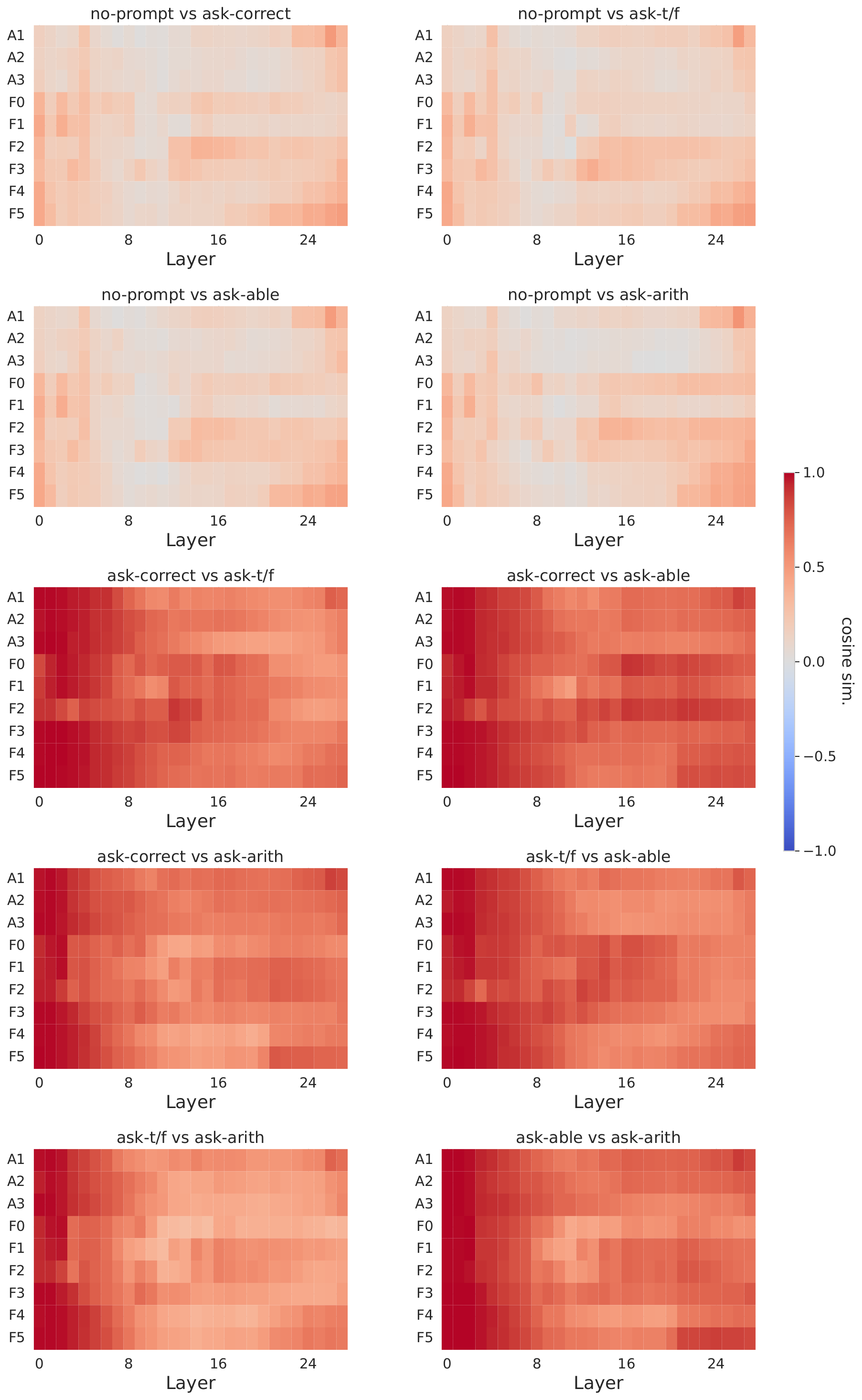}
    \caption{Cosine similarity of truth probes for \textbf{Llama-3.2-3B-Instruct}.}
\end{figure*}

\begin{figure*}[h]
    \centering
    \includegraphics[width=\linewidth]{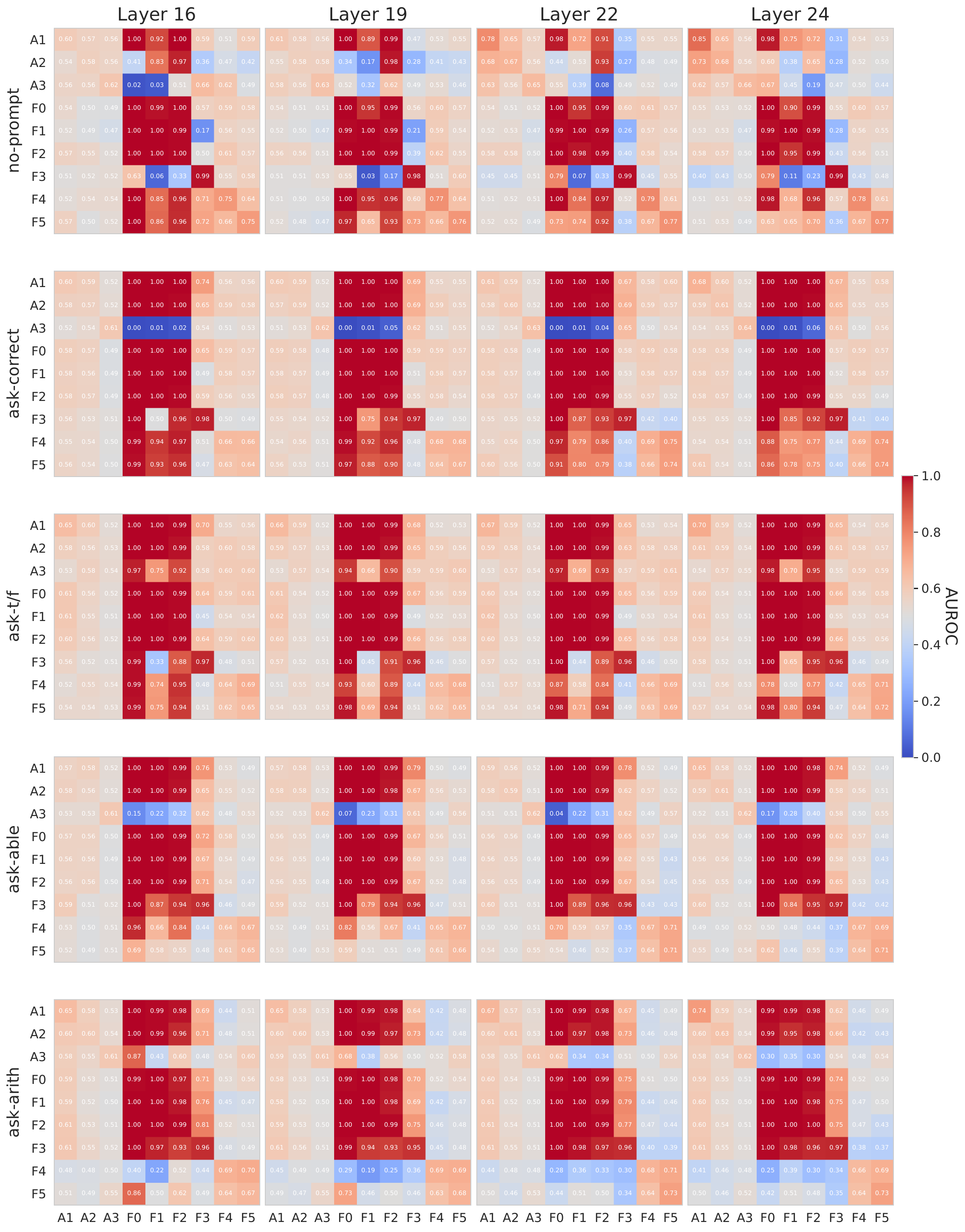}
    \caption{Cross-task generalization heatmaps for \textbf{Llama-3.2-3B-Instruct}.}
\end{figure*}

\begin{figure*}[h]
\centering
{\includegraphics[width=\linewidth]{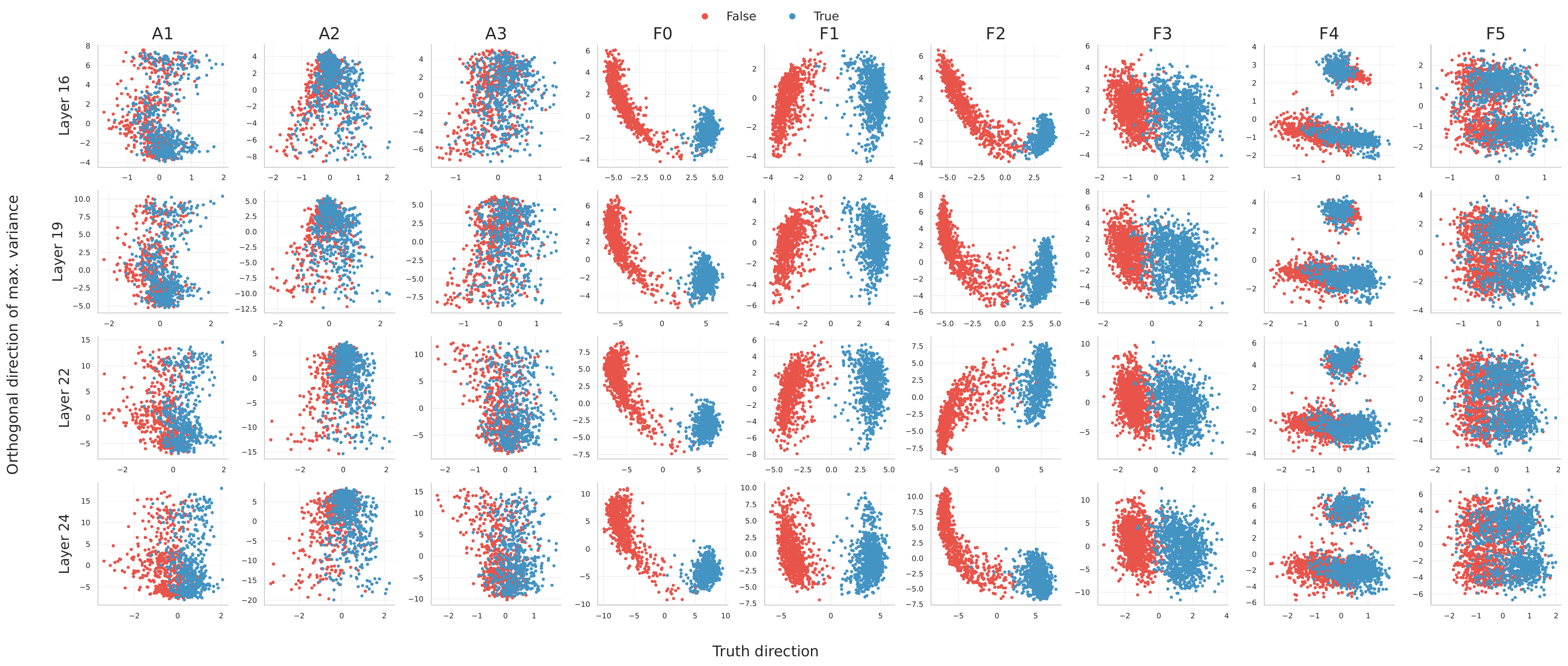}
}
\caption{Activation projections for \textbf{Llama-3.2-3B-Instruct}.}
\end{figure*}

\subsection{Gemma-2-2b-it}

\begin{figure*}[h]
\centering
\begin{minipage}[h]{0.42\textwidth}
    \vspace{0pt}
    \subfloat[In-domain test AUROC across layers.]{
        \includegraphics[width=.95\linewidth]{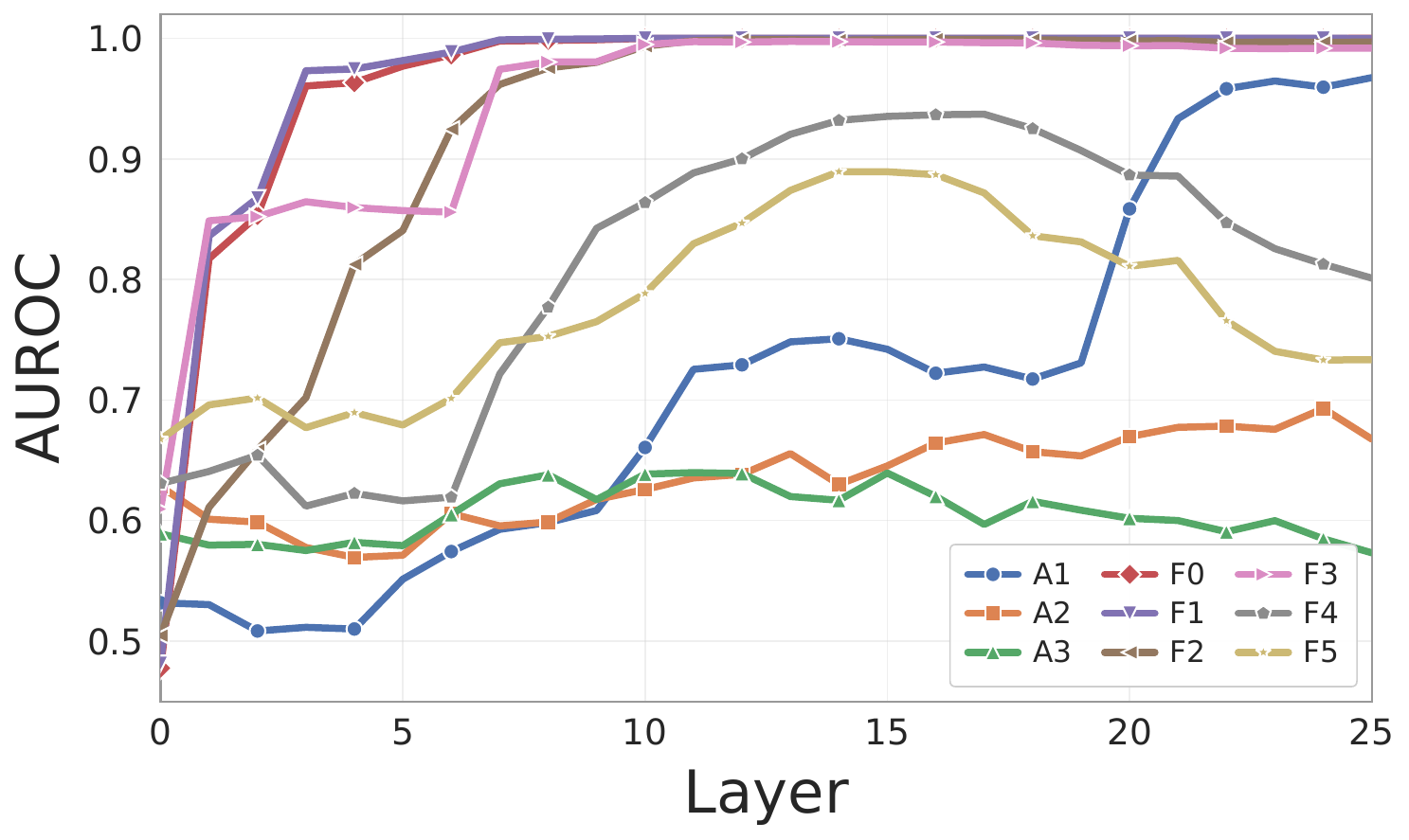}%
    }\\[0.5em]
    \subfloat[F0-trained probe cross-task generalization across layers.]{
        \includegraphics[width=.95\linewidth]{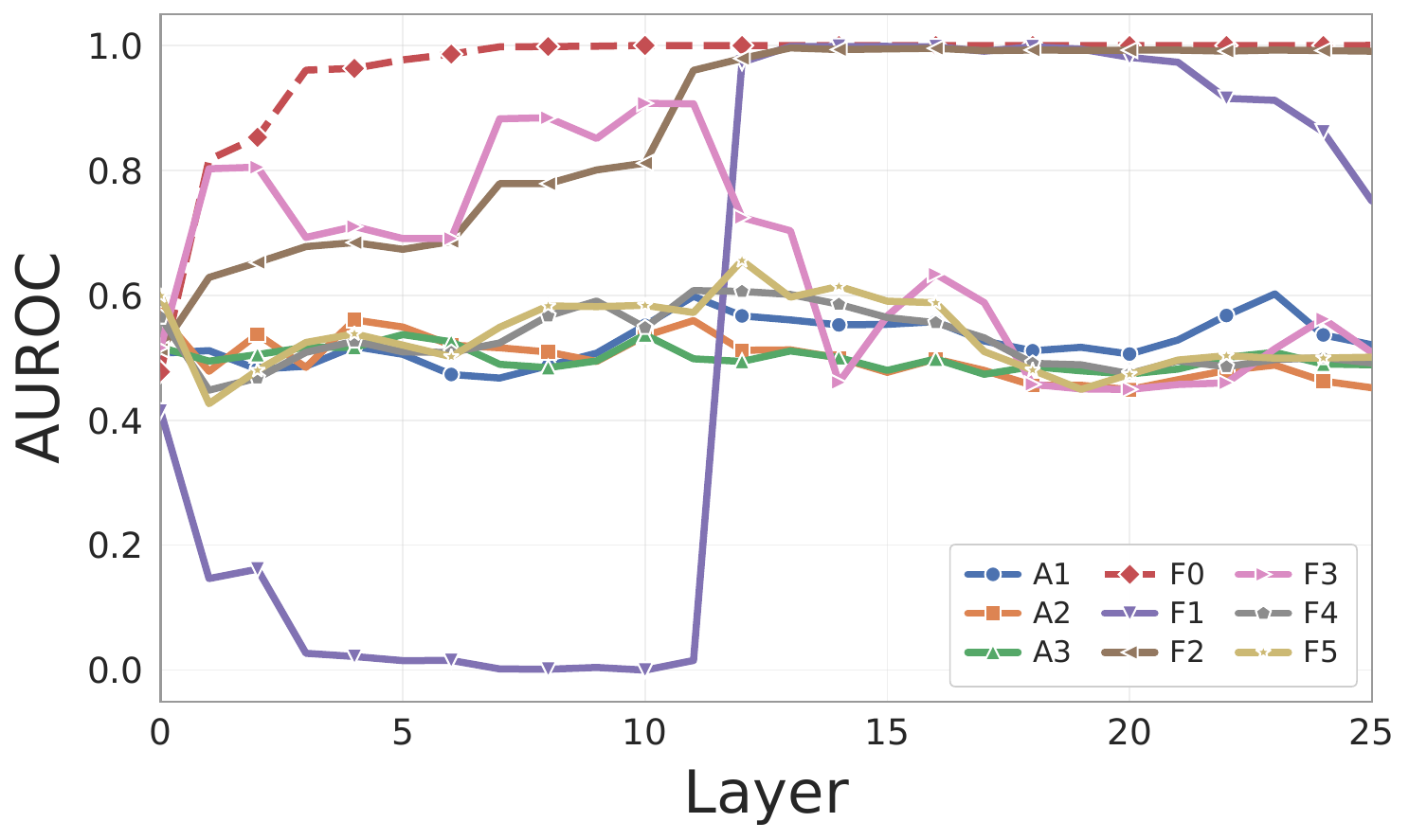}
    }
\end{minipage}
\hfill
\begin{minipage}[h]{0.57\textwidth}
    \vspace{0pt}
    \subfloat[Cosine similarity of probes across all pairs of layers for each task.]{
        \includegraphics[width=\linewidth]{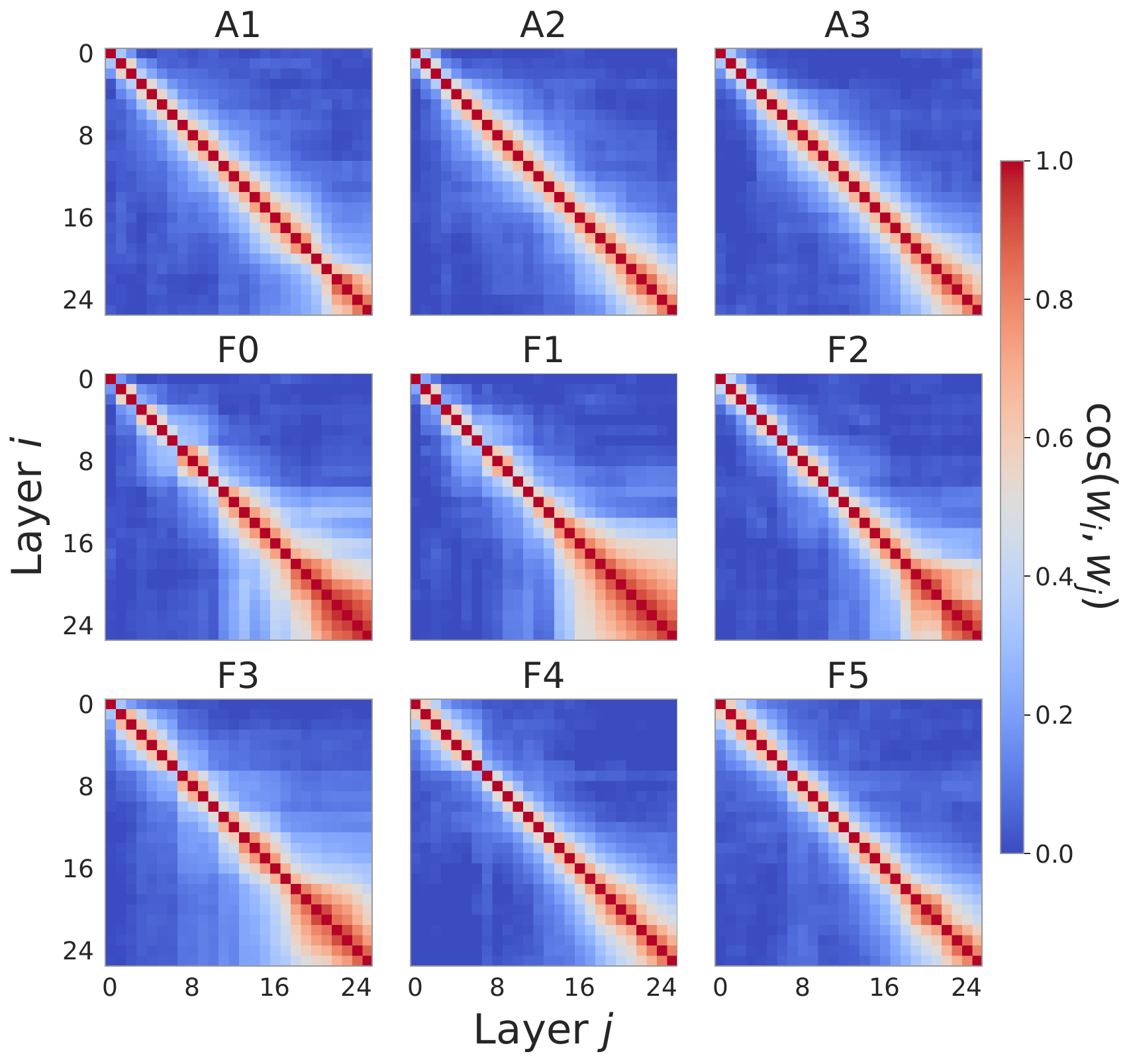}%
    }
\end{minipage}
\caption{Layer dependence of truth directions for \textbf{Gemma-2-2b-it}. (a) In-domain test-set performance. (b) F0-probe cross-task generalization. (c) Cosine similarity of probes across layer pairs.}
\label{app:fig:layer_dependence_combined_gemma2b}
\end{figure*}

\begin{figure*}
    \centering
    \includegraphics[width=.9\linewidth]{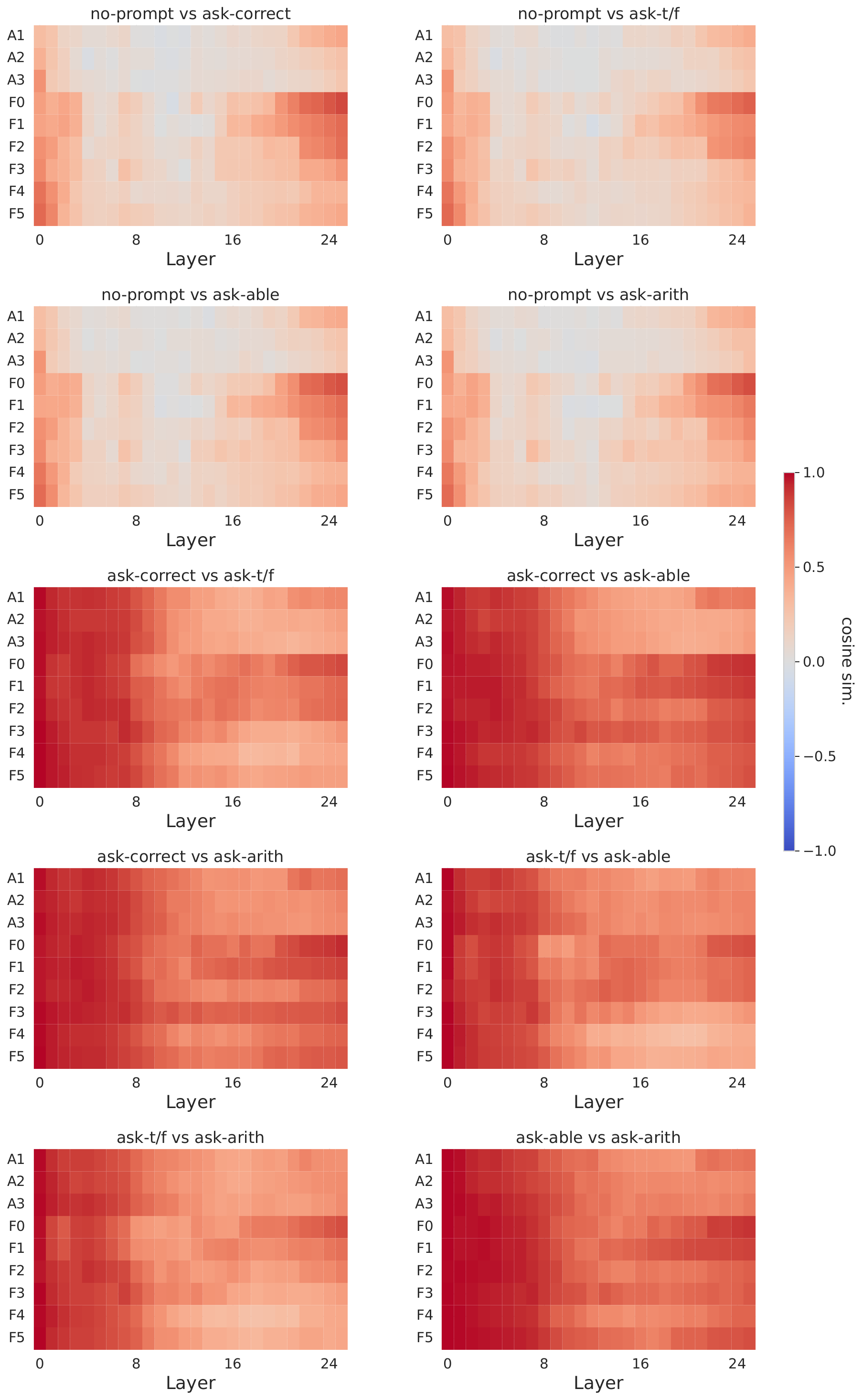}
    \caption{Cosine similarity of truth probes for \textbf{Gemma-2-2b-it}.}
\end{figure*}

\begin{figure*}
    \centering
    \includegraphics[width=\linewidth]{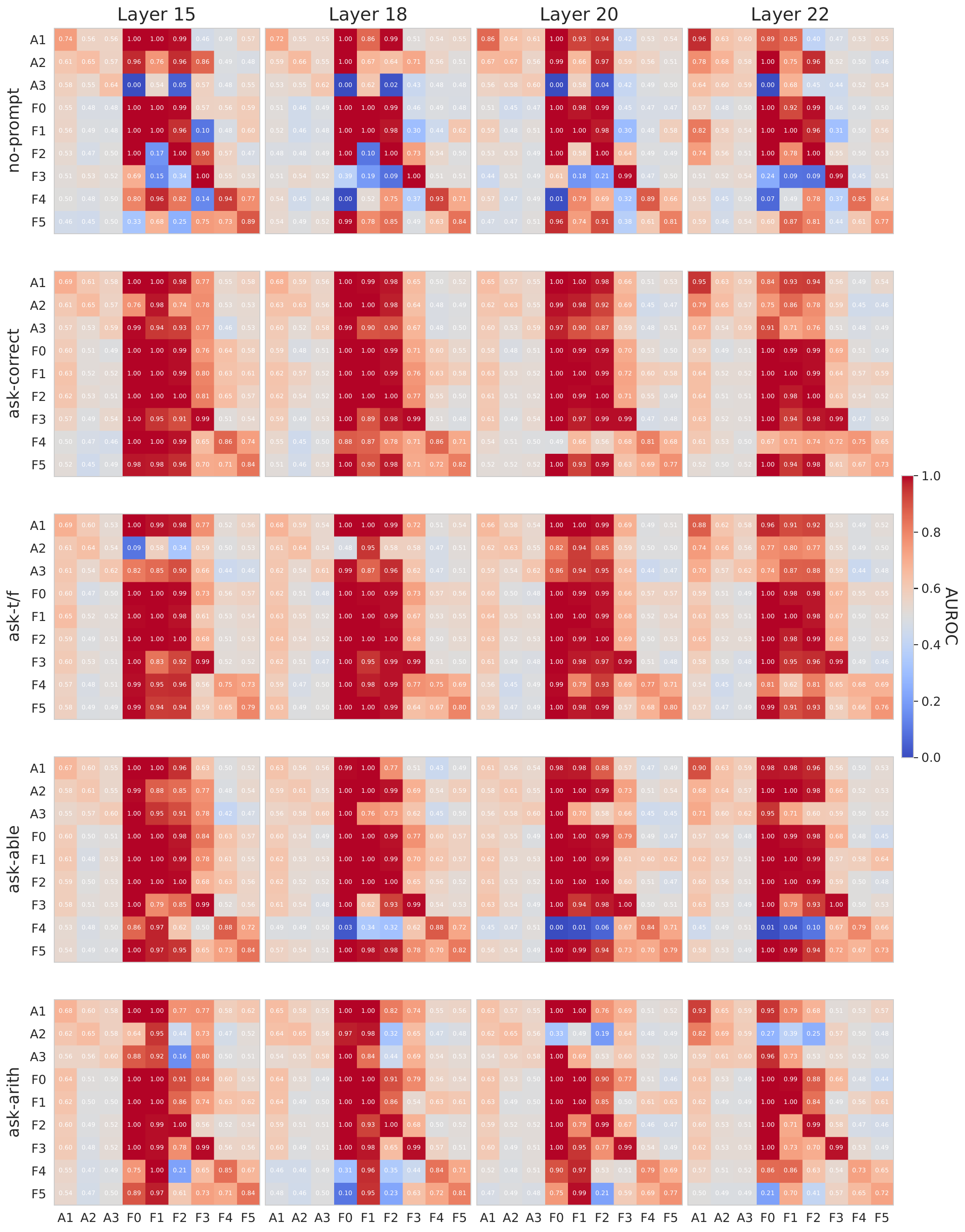}
    \caption{Cross-task generalization heatmaps for \textbf{Gemma-2-2b-it}.}
\end{figure*}

\begin{figure*}
    \centering
    \includegraphics[width=\linewidth]{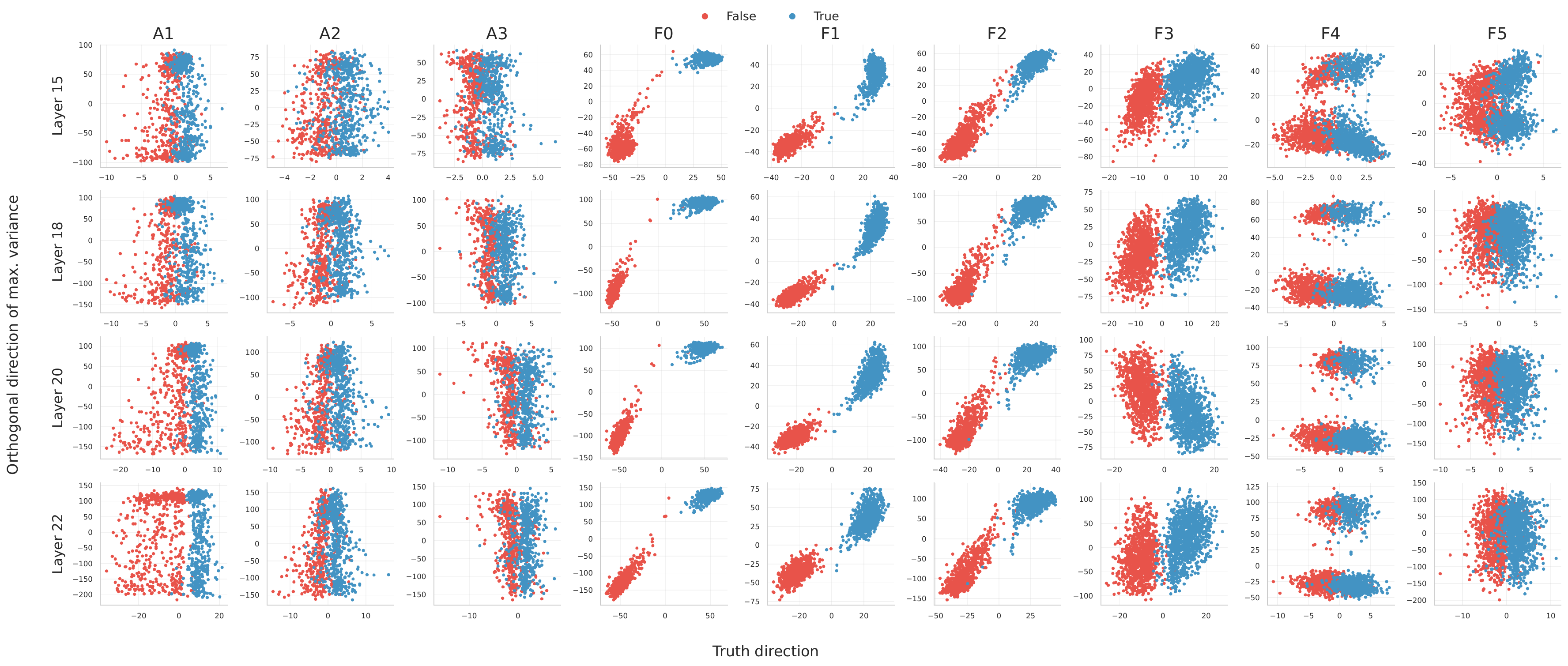}
    \caption{Activation projections for \textbf{Gemma-2-2b-it}.}
\end{figure*}

\subsection{Gemma-2-9b-it}

\begin{figure*}[h]
\centering
\begin{minipage}[h]{0.42\textwidth}
    \vspace{0pt}
    \subfloat[In-domain test AUROC across layers.]{
        \includegraphics[width=.95\linewidth]{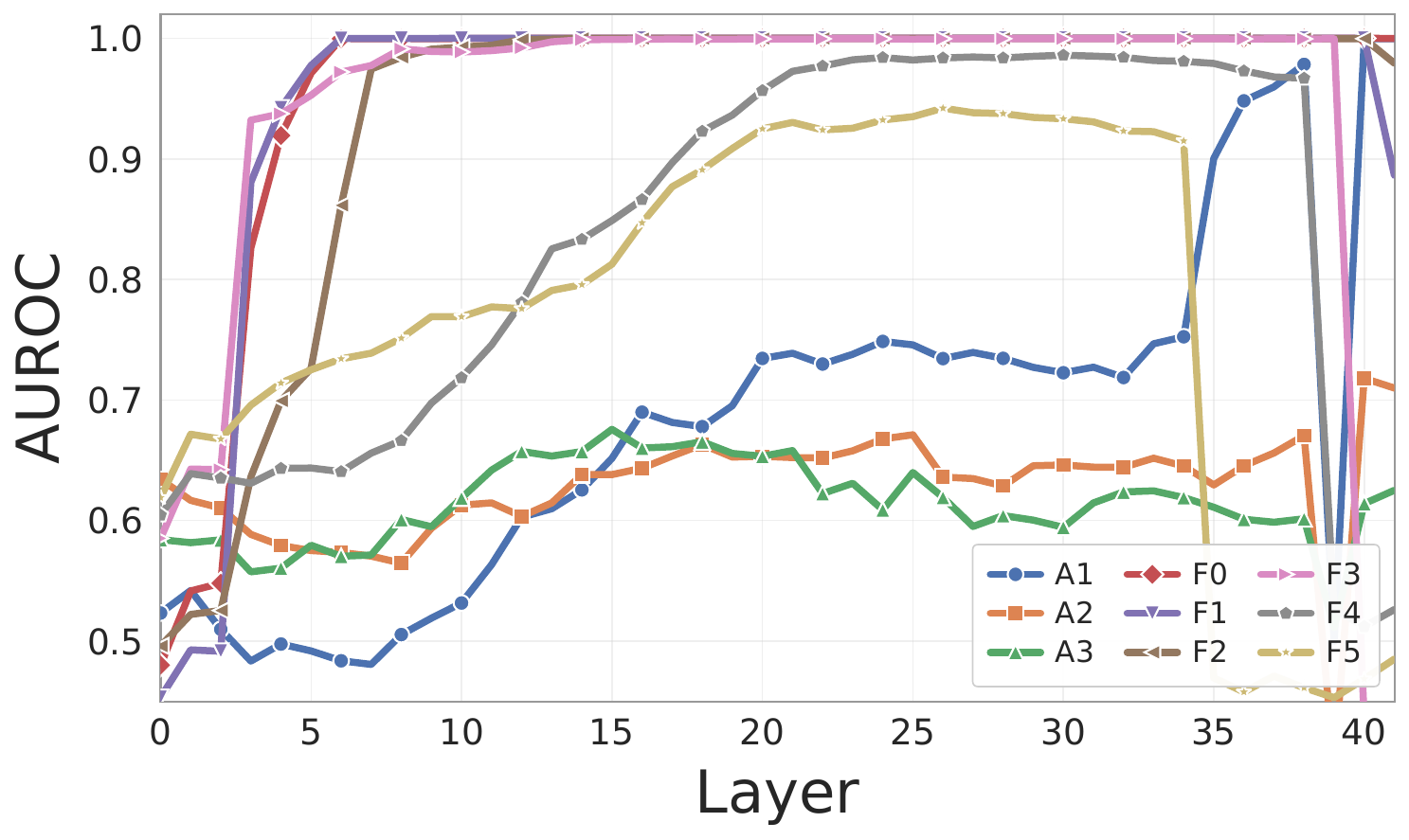}
    }\\[0.5em]
    \subfloat[F0-trained probe cross-task generalization across layers.]{
        \includegraphics[width=.95\linewidth]{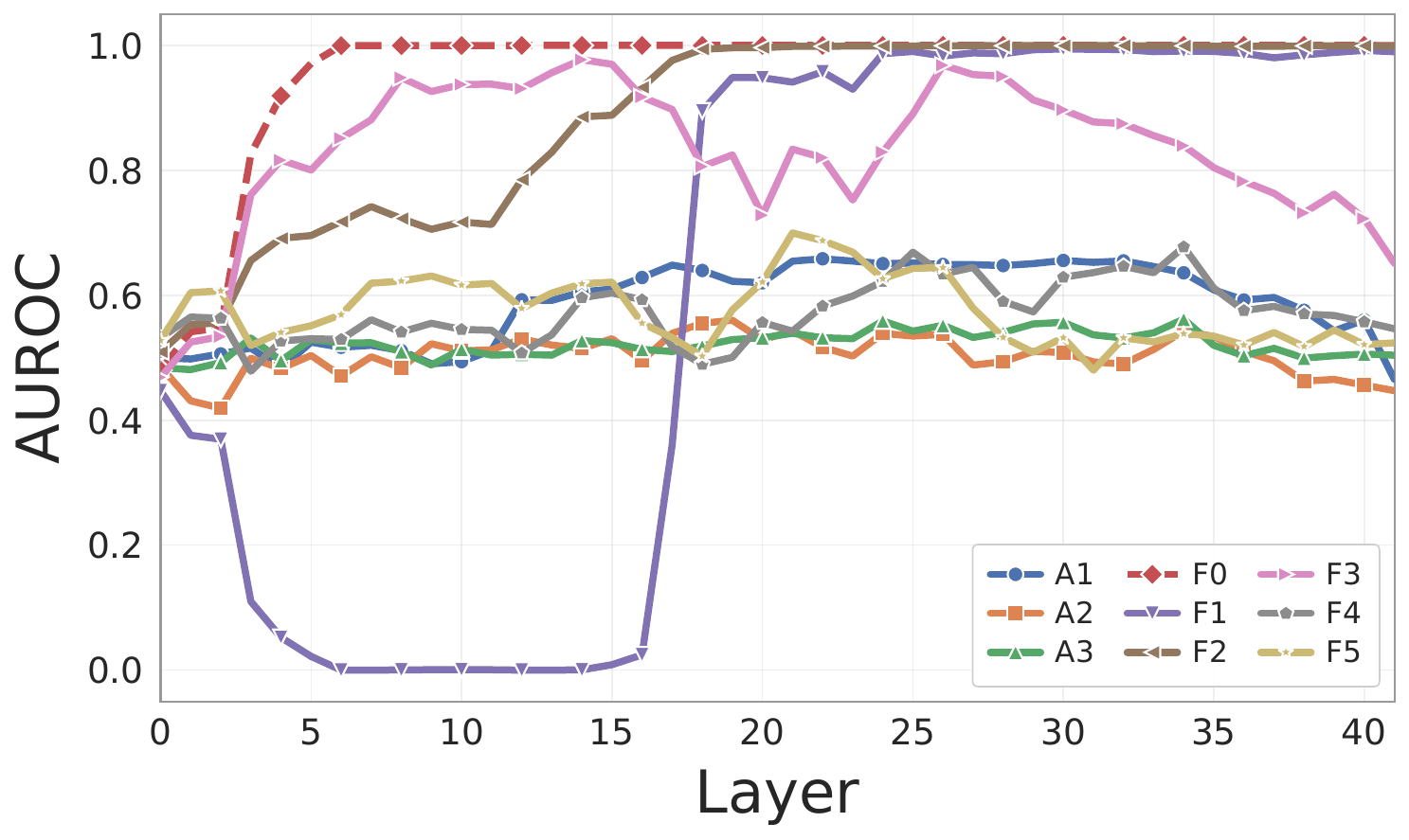}
    }
\end{minipage}
\hfill
\begin{minipage}[h]{0.56\textwidth}
    \vspace{0pt}
    \subfloat[Cosine similarity of probes across all pairs of layers for each task.]{
        \includegraphics[width=0.99\linewidth]{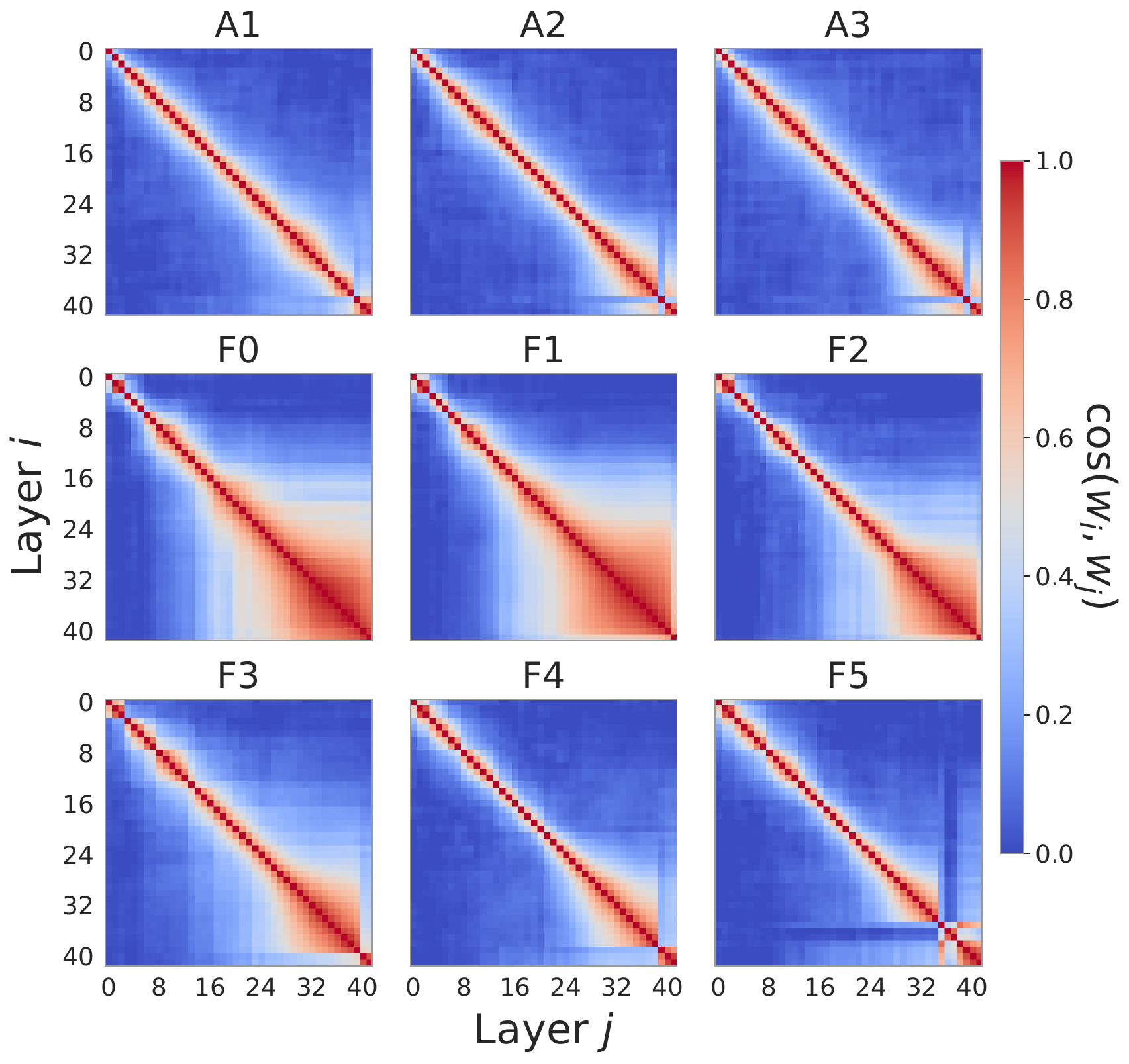}%
    }
\end{minipage}
\caption{Layer dependence of truth directions for \textbf{Gemma-2-9b-it}. (a) In-domain test-set performance. (b) F0-probe cross-task generalization. (c) Cosine similarity of probes across layer pairs.}
\label{app:fig:layer_dependence_combined_gemma9b}
\end{figure*}

\begin{figure*}
    \centering
    \includegraphics[width=.9\linewidth]{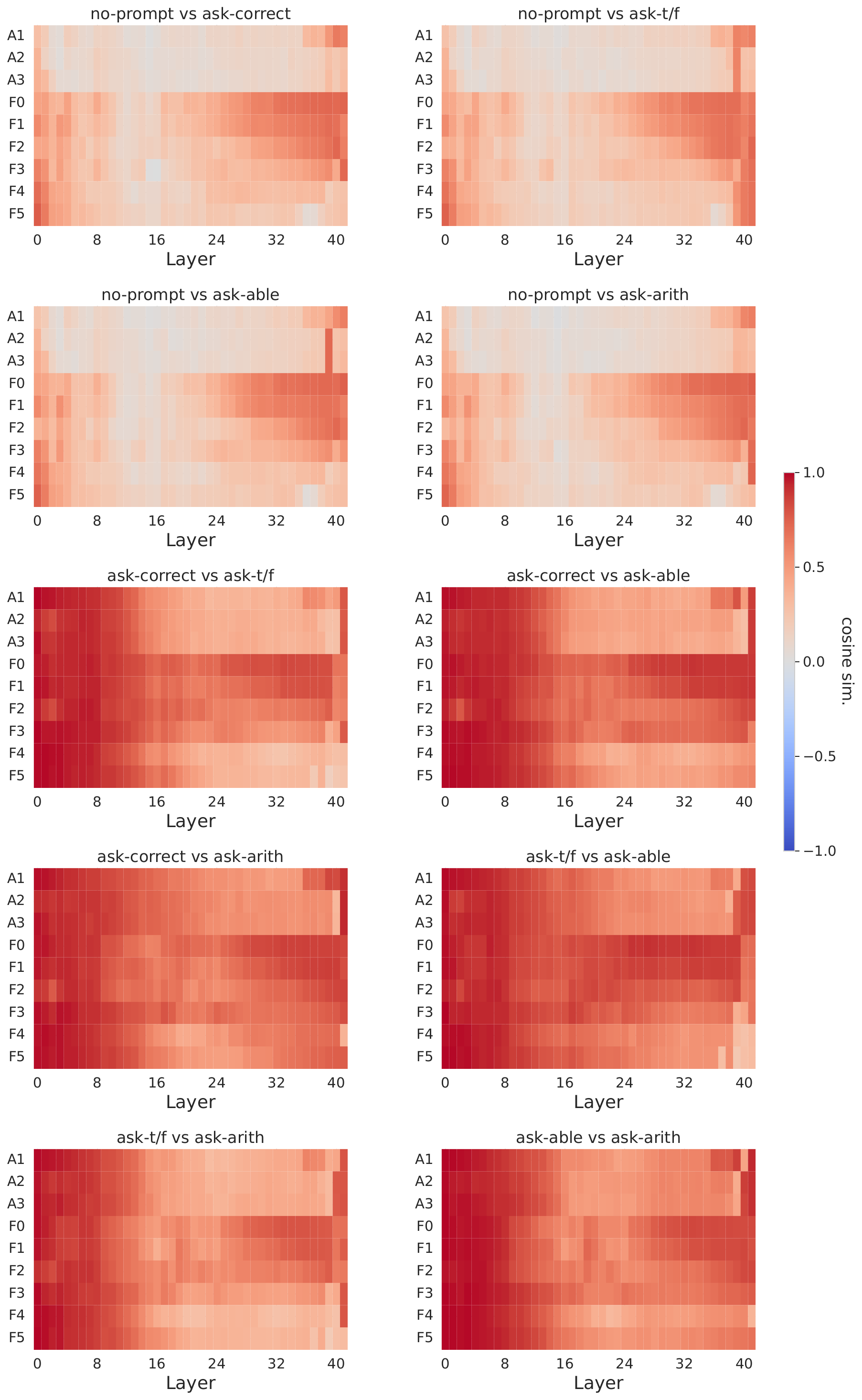}
    \caption{Cosine similarity of truth probes for \textbf{Gemma-2-9b-it}.}
\end{figure*}

\begin{figure*}
    \centering
    \includegraphics[width=\linewidth]{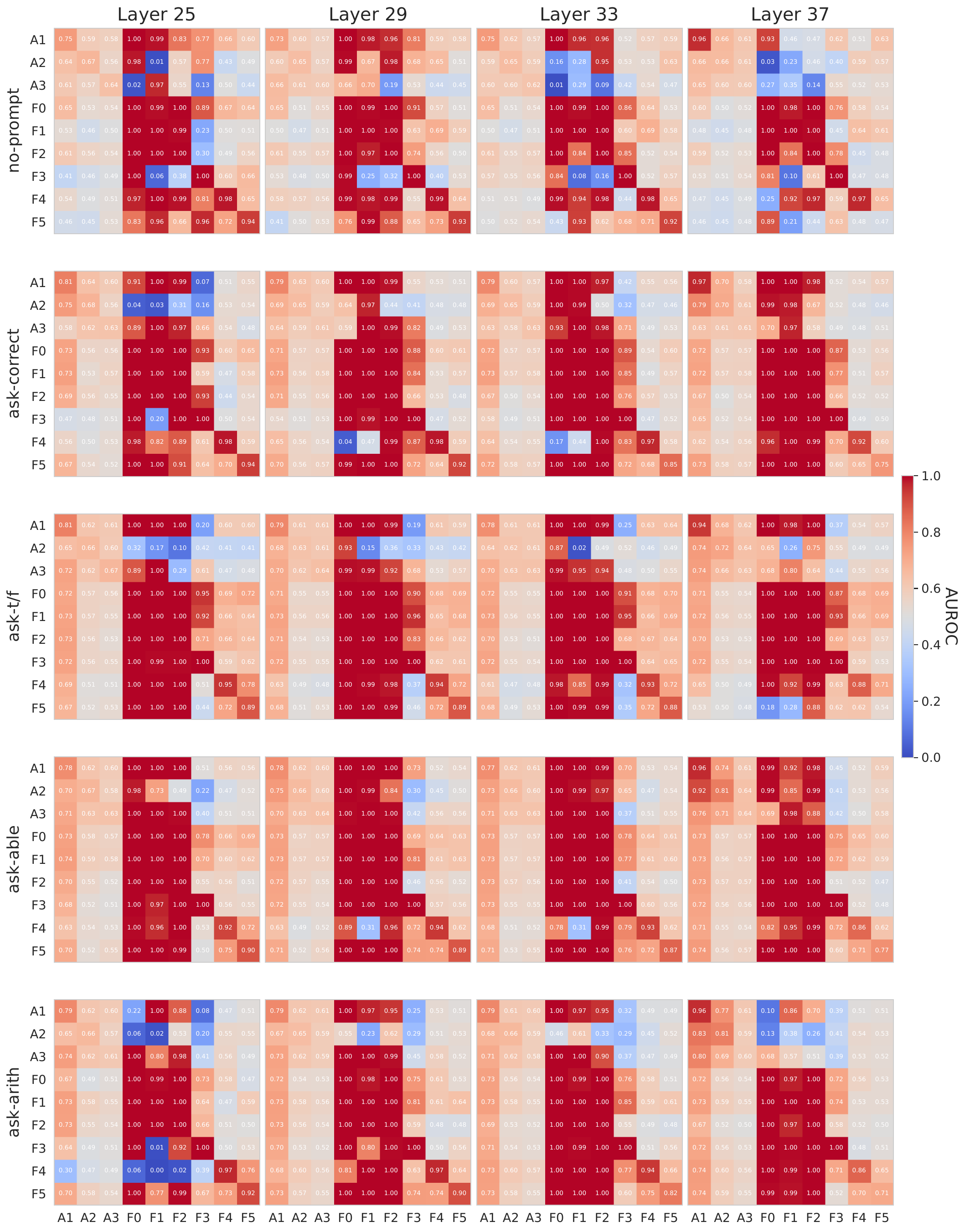}
    \caption{Cross-task generalization heatmaps for \textbf{Gemma-2-9b-it}.}
\end{figure*}

\begin{figure*}[t]
\centering
{\includegraphics[width=\linewidth]{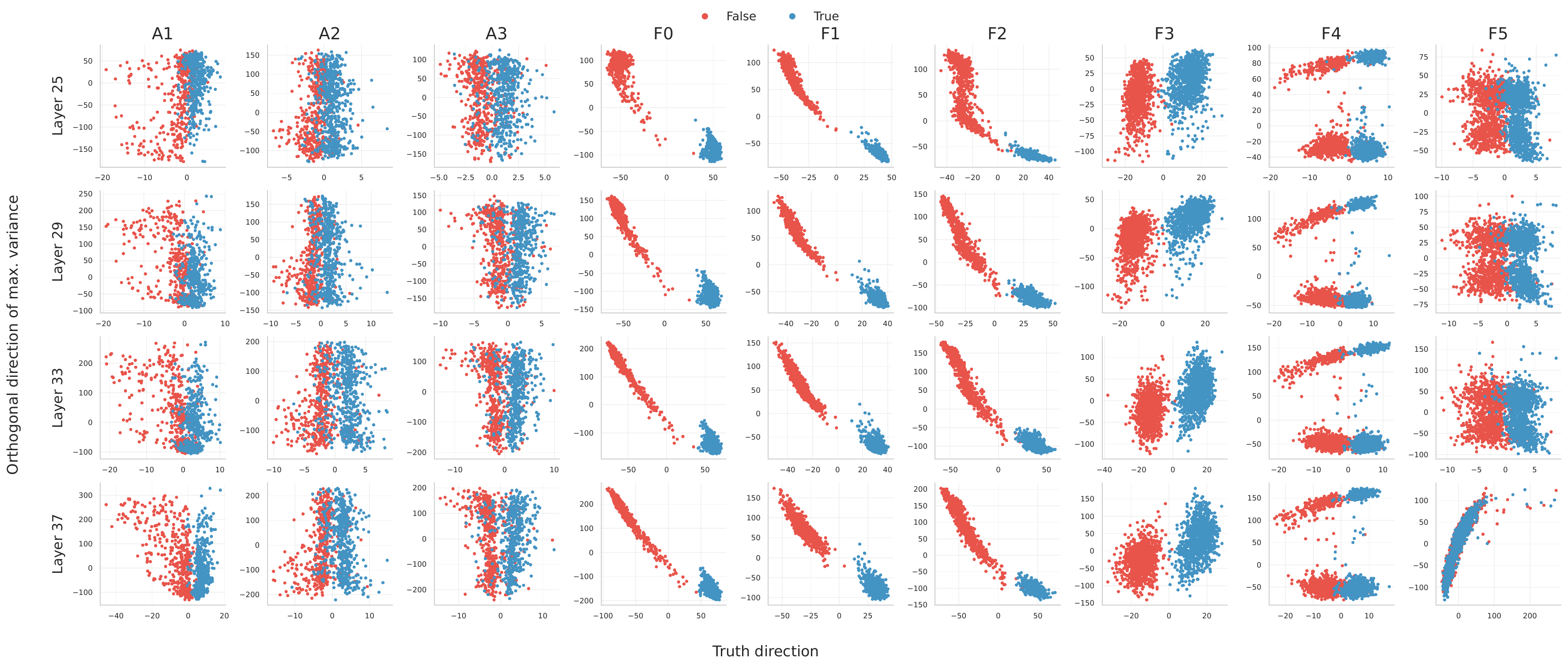}
}
\caption{Activation projections for \textbf{Gemma-2-9b-it}.}
\end{figure*}

\end{document}